\documentclass[twopage]{article}

\usepackage[accepted]{aistats2021}
\usepackage[T1]{fontenc}
\usepackage{tikz}
\usetikzlibrary{arrows}
\usetikzlibrary{calc}
\usetikzlibrary{snakes}
\usepackage{cite}
\usepackage{graphicx}
\usepackage{array}
\usepackage{mdwmath}
\usepackage{mdwtab}
\usepackage{eqparbox}
\usepackage{fixltx2e}
\usepackage{url}
\usepackage{pgf}
\usepackage{lipsum}
\usepackage{multirow}
\usepackage{subcaption}
\usepackage{amsfonts,amssymb,amsthm}
\usepackage{xspace}
\usepackage{xcolor}
\usepackage{bm}
\usepackage{bbm}
\usepackage{booktabs}
\usepackage{etoolbox}
\usepackage{makecell}
\usepackage{nicefrac}
\usepackage{textcomp}
\usepackage{stmaryrd}
\usepackage{mathrsfs}
\usepackage{paralist}
\usepackage{comment}

\usepackage[font=footnotesize,labelsep=space,labelfont=bf]{caption}

\usepackage{url}

\usepackage[pdftex,bookmarks=true,pdfstartview=FitH,colorlinks,linkcolor=black,filecolor=black,citecolor=black,urlcolor=black,pagebackref=false]{hyperref}
\makeatletter

\newcommand{\Rmnum}[1]{\expandafter\@slowromancap\romannumeral #1@}
\makeatother

\newtheorem*{theorem*}{Theorem}
\newtheorem*{lemma*}{Lemma}
\newtheorem*{cor*}{Corollary}

\newcommand{\namedref}[2]{\hyperref[#2]{#1~\ref*{#2}}}

\newcommand{\Sectionref}[1]{\namedref{Section}{sec:#1}}
\newcommand{\Subsectionref}[1]{\namedref{Section}{subsec:#1}}
\newcommand{\Subsubsectionref}[1]{\namedref{Section}{subsubsec:#1}}
\newcommand{\Appendixref}[1]{\namedref{Appendix}{app:#1}}

\newcommand{\Figureref}[1]{\namedref{Figure}{fig:#1}}

\newcommand{\Pageref}[1]{\hyperref[#1]{page~\pageref*{#1}}}

\definecolor{darkred}{rgb}{0.5, 0, 0} 
\definecolor{darkblue}{rgb}{0,0,0.5} 
\definecolor{darkblack}{rgb}{0,0,0} 

\hypersetup{
    colorlinks=true,
    linkcolor=darkblack,
    citecolor=darkblack,
    urlcolor=darkblack   
}

\newcommand{\bZ}{\ensuremath{{\bf Z}}\xspace}

\newcommand{\bX}{\ensuremath{{\bf X}}\xspace}
\newcommand{\bN}{\ensuremath{{\bf N}}\xspace}

\newcommand{\bbE}{\ensuremath{\mathbb{E}}\xspace}

\newcommand{\I}{\ensuremath{\mathcal{I}}\xspace}

\newcommand{\E}{\ensuremath{\mathcal{E}}\xspace}

\renewcommand{\paragraph}[1]{\smallskip\noindent{\bf #1}~}

\newcommand\independent{\protect\mathpalette{\protect\independenT}{\perp}}
\def\independenT#1#2{\mathrel{\rlap{$#1#2$}\mkern2mu{#1#2}}}

\usepackage{algorithm}
\usepackage{algorithmicx}
\usepackage{algpseudocode}
\usepackage{graphicx}
\usepackage{grffile}
\algdef{SE}[DOWHILE]{Do}{doWhile}{\algorithmicdo}[1]{\algorithmicwhile\ #1}%
\usepackage{authblk}

\usepackage{dcolumn}
\usepackage{booktabs}
\usepackage{tikz}
\usetikzlibrary{fit, positioning,shapes,arrows}

\newcolumntype{M}[1]{D{.}{.}{1.#1}}

\usepackage{etoolbox}\AtBeginEnvironment{algorithmic}{\small}

\begin{document}

%

%

\twocolumn[

\aistatstitle{On Shapley Credit Allocation for Interpretability}

\aistatsauthor{ Debraj Basu}

\aistatsaddress{ Adobe Inc, San Jose\\dbasu@adobe.com} ]

\begin{abstract}
We emphasize the importance of asking the right question when interpreting the decisions of a learning model. We discuss a natural extension of the theoretical machinery from Janzing et. al. 2020, which answers the question ``Why did my model predict a person has cancer?'' for answering a more involved question, ``What caused my model to predict a person has cancer?'' While the former quantifies the direct effects of variables on the model, the latter also accounts for indirect effects, thereby providing meaningful insights wherever human beings can reason in terms of cause and effect. We propose three broad categories for interpretations: \textit{observational}, \textit{model-specific} and \textit{causal} each of which are significant in their own right. Furthermore, this paper quantifies feature relevance by weaving different natures of interpretations together with different measures as characteristic functions for Shapley symmetrization. Besides the widely used expected value of the model, we also discuss measures of statistical uncertainty and dispersion as informative candidates, and their merits in generating explanations for each data point, some of which are used in this context for the first time. These measures are not only useful for studying the influence of variables on the model output, but also on the predictive performance of the model, and for that we propose relevant characteristic functions that are also used for the first~time. 
\end{abstract}

\section{Introduction}
\label{sec:intro}
Non-linear models with large degrees of freedom have become the workhorse of modern industrial automation.
 The lack of transparency together with the need for fairness and accountability underscores the importance of a clear understanding of their functioning \cite{lipton} which is widely referred to as \textit{model interpretability}. Our primary focus is on explaining model decisions for every data point. Although the sentiment of advocating transparency and thereby fairness or identifying the lack thereof is clear, what is left vague is the very definition of what interpretability must entail \cite{lipton}.

Among many developments in the recent past \cite{deeplift,lime,ig,igor,lundberg,lundberg_shap,survey}, an additive feature attribution method known as SHAP \cite{lundberg_shap} has emerged as the state-of-the-art, by leveraging a solution concept from cooperative game theory known as Shapley values \cite{shapley}, bringing several theoretical benefits which translate into significantly powerful empirical results in common tasks designed for evaluating model interpretability including their resonance with a human audience. By successfully navigating complex non-linear relationships between features and the model output, Shapley values have become the de facto standard for assigning feature relevance towards the decisions made by a learning model. 
The Shapley symmetrization provides a meaningful summarization of the marginal effects of a variable on the model output when in conjunction with different combinations of other variables. However, it is only a means of fairly distributing the generated surplus, between different players of a game. 
In its current form where it goes by the name of SHAP\cite{lundberg_shap}, it exhibits certain finer drawbacks which are discussed in \cite{shap_problems, janzing-causal,ma-shap,najmi}.

Perhaps, these scores have certain \textit{observational} implications which are not just salient to the model but also capture underlying relational dependencies between the variables which are governed by the data, resulting in irrelevant variables surfacing as important. \cite{janzing-causal} shows that both these notions can be separated by using a different characteristic function based on a \textit{causal} reconstruction which only accounts for the direct effects of a variable on the model output. We extend this to also account for indirect \textit{causal} effects of variables, which could possibly result in a variable not used by a model surfacing as important. We suggest three broad categories for the natures of interpretation: \textit{observational} \cite{lundberg_shap}, \textit{model-specific} \cite{janzing-causal} and \textit{causal}~\Sectionref{noi}.

This distinction between the nature of different kinds of interpretations involves extensive use of Pearl's $do$ operator \cite{pearl_2009}, for deriving feature relevance with significantly diverse implications. Furthermore, we show that these implications can be perceived through various metrics such as the well known expected value, as well as those of statistical uncertainty and dispersion. The nature of interpretation we seek, together with the relevant measure is what interpretability entails. We show how this combined framework can be used for studying the effects of variables on either the model output or the predictive performance of a model.

\textbf{Related Work.} In his seminal work on cooperative games \cite{shapley}, Lloyd Shapley introduced a solution for distributing the total gains to all collaborating players. \cite{igor} discussed sampling-based approximations for attributing feature relevance in machine learning.

Our work is most closely related to \cite{janzing-causal,lundberg_shap,janzing-cic,pearl_2009,klein-1,klein-2}. \cite{lundberg_shap} used Shapley values to unify many different methods from the literature and demonstrated the possibility of more efficiently computing Shapley values using an equivalent weighted linear regression from \cite{charnes1988}.
\cite{janzing-causal} argued that while the intention of \cite{lundberg_shap} is well placed, their proposition surfaces irrelevant variables as important. \cite{janzing-causal} suggests a small technical adjustment for fixing this which demonstrates that model interpretability is a \textit{causal} problem. 

Our work stresses on our responsibility being to identify the relevant nature of interpretation being sought before we apply one of three different types namely \textit{observational}, \textit{model-specific} and \textit{causal} (see \Sectionref{noi}). We show that the first two which are preexisting notions from \cite{lundberg_shap} and \cite{janzing-causal}, naturally extend to \textit{causal} interpretations which leverages Pearl's $do$ operator \cite{pearl_2009} as well as the underlying \textit{causal} graph. 

In \Sectionref{moi} we combine different natures of interpretation, together with different measures of interpretation, some of which have been studied at a global level such as \textit{causal} information contribution (CIC) and \textit{causal} variance contribution (CVC) \cite{janzing-cic}, for generating meaningful explanations for each data point. We also demonstrate the benefits of cumulative pairwise Shannon entropy \cite{klein-1,klein-2} as another candidate measure of interpretation drawing analogies with Shannon entropy and variance.
We apply these concepts for attributing relevance to variables not only towards the model output, but also the predictive performance (loss) of the model which we show relies on the generic definition of Shapley values, and a \textit{causal} graph of all variables including the target, model and the loss.

\section{Shapley Values}
\label{sec:shapley}
Throughout our discussion we use the following notation: $\textbf{X}=\{X_1,X_2,\ldots,X_p\}$ denotes all variables, and $Y$ denotes the target variable which is of interest to a predictive model. $\Pr(\bX,Y)$ denotes the joint distribution over all variables. Let $S$ be a subset of $P=[p]:=\{1,2,\ldots,p\}$, then $\bX_S=\{X_i:i\in S\}$.

A coalitional game is defined by a tuple $\langle\bN,v\rangle$ where $\bN=[n]$ is a finite set of $n$ players, and $v:2^\bN\rightarrow\mathbb{R}$ is a characteristic function such that $v(\phi)=0$. $v$ determines the value of each subset of \bN. Shapley values provide a unique and fair allocation of the value $v(\bN)$ to each player $i\in\bN$, the payoff being denoted by $\phi_i(v)$ satisfies several properties that are desirable of distributed payoffs as seen in \cite{lundberg_shap,ma-shap}.
The analytical form of the Shapley value given by, \vspace{-2mm}
\begin{align*}
\phi_i(v)=\frac{1}{|\bN|}\sum_{S\subseteq\bN-\{i\}}\binom{|\bN|-1}{|S|}^{-1}\left[v(S \cup \{i\})-v(S)\right].
\end{align*}
represents the weighted average of the influence of $i$ in each coalition, denoted by $\phi_i^S(v):=v(S \cup \{i\})-v(S)$.

Now consider a predictive model $g(\bX)$ for the target of interest $Y$. For a characteristic function $m:2^{\bX}\rightarrow \mathbb{R}$, the corresponding co-operative game is given by $\langle\bX,m\rangle$, where $m$ maps a subset of $\bX$ to a real number representing its contribution to the output of model $g$. 
A well known candidate for $m$ is $\bbE[g(\bX)|\bX_S=x_s]-\bbE[g(\bX)]$. Here $m=g(x)-\bbE[g(\bX)]$ when $S=P$ and $0$ when $S=\emptyset$. A game theoretic view is intuitive here, because all $P$ players are working together to generate a surplus of $g(x)-\bbE[g(\bX)]$. For variables observed in the following order $X_1,X_2,\ldots,X_p$, we have \vspace{-2mm}
\begin{small}\begin{align*}&g(x)-\bbE[g(\bX)]\\
&=\bbE[g(\bX)|\bX=x]-\bbE[g(\bX)|\bX_{[p-1]}=x_{[p-1]}]\\
&+\bbE[g(\bX)|\bX_{[p-1]}=x_{[p-1]}]-\bbE[g(\bX)|\bX_{[p-2]}=x_{[p-2]}]\\
&+\ldots+\bbE[g(\bX)|\bX_1=x_1]-\bbE[g(\bX)].\end{align*} \end{small}
Therefore in this ordering, the marginal contribution of $x_i$ is given by $\bbE[g(\bX)|\bX_{[i]}=x_{[i]}]-\bbE[g(\bX)|\bX_{[i-1]}=x_{[i-1]}]$. By accounting for all possible orderings which correspond to different ``paths'' to generating the surplus, we arrive at Shapley values as a unique solution.  

\section{Nature of Interpretation}
\label{sec:noi}
%
In this section, we understand the differences in the natures of interpretation and the corresponding analysis that entails each one of them. This refines our expectations from model interpretability through the realization that understanding model response is not just an analysis of the internals of a model, but must also account for the dependencies between variables. In the \textit{causal} graphical model below, we have four indicator random variables: genotype ($X_1$), smoking ($X_2$), lung cancer ($Y$), and chest pain ($X_3$). The following relationships hold, \textit{(i)} No two variables are independent of each other; \textit{(ii)} Chest pain is independent of both genotype and smoking, given lung cancer; \textit{(iii)} Chest pain cannot cause lung cancer; \textit{(iv)} Smoking is a direct cause of lung cancer; \textit{(v)} Genotype has a direct effect on lung cancer and can also cause a person to smoke or not which in turn affects lung cancer. 
\begin{small}
\begin{tikzpicture}[
  node distance=0.4cm and 1.5cm,
  mynode/.style={draw,circle,text width=0.5cm,align=center}
]
\node[mynode] (gt) {$X_1$};
\node[mynode,below right=of gt] (lc) {$Y$};
\node[mynode,below left=of gt] (sm) {$X_2$};
\node[mynode,right=of lc] (cp) {$X_3$};
\path (gt) edge[-latex] (lc)
(gt) edge[-latex] (sm) 
(sm) edge[-latex] (lc)
(lc) edge[-latex] (cp);
\end{tikzpicture}
\end{small}

We find that whenever human beings can reason in terms of cause and effect relationships, \textit{observational} insights can be misleading \cite{shap_problems,janzing-causal}. In such a scenario, it is prudent to first settle on the nature of interpretations sought between \textit{model-specific} and \textit{causal}, before we look for answers. \Appendixref{num} discusses a simple example distinguishing between the three different types of interpretations discussed in this section.

\textbf{Observational.} Assume that the variables follow a \textit{causal} Bayesian network given by edges $X_1\rightarrow X_2$, $X_1\rightarrow Y$ and $Y\rightarrow X_3$. First we will employ $\mathbb{E}\left[g(\bX)|\bX_S=x_S\right]-\bbE[g(\bX)]$ as $m$ as was as discussed in \Sectionref{shapley}, where $g$ is a function of $X_1$ and $X_3$, and independent of all other variables given $X_1$ and $X_3$. Therefore we have $X_1\rightarrow g$ and $X_3\rightarrow g$. 
 \begin{small}
\begin{tikzpicture}[
  node distance=0.4cm and 1.3cm,
  mynode/.style={draw,circle,text width=0.5cm,align=center}
]
\node[mynode] (gt) {$X_1$};
\node[mynode,right=of gt] (lc) {$Y$};
\node[mynode,left=of gt] (sm) {$X_2$};
\node[mynode,right=of lc] (cp) {$X_3$};
\node[mynode,below=of cp] (g) {$g$};
\path (gt) edge[-latex] (lc)
(gt) edge[-latex] (sm)
(gt) edge[-latex] (g)
(cp) edge[-latex] (g)
(lc) edge[-latex] (cp);

\end{tikzpicture} \end{small}

Now on computing Shapley values for a data point $x$, we may observe a non zero $\phi_1(m)$ and $\phi_3(m)$, which denote relative contributions of $X_1$ and $X_3$ towards $g$. This should not be misunderstood as the effect of changing or masking these variables on $g$.

We consolidate this argument by computing Shapley values for $\langle\{X_1,X_2,X_3\}, m\rangle$.
We know that the model output is d-separated from smoking ($X_2$) when the gene ($X_1$) is present. However, the values paint a different picture because smoking ($X_2$) gives an appearance of having a certain influence on the predictor for cancer ($g$) via a backdoor path which could possibly surface $X_2$ as important. 
Therefore the characteristic function ($m$) here merely captures associative relations between variables and the model output. Further more the $\phi_1(m)$ and $\phi_3(m)$ may not be same as earlier, and will be adjusted to satisfy $\phi_1(m) + \phi_2(m) + \phi_3(m)=m(x_1, x_2,x_3)$.
%
%

Therefore when we examine the directionality of the relationships between variables, we find that the measure of influence using Shapley values can be misleading in cases where we are looking for \textit{causal} explanations. 
However, note that the cause of the mismatch with our expectations is not the concept of Shapley values, but rather the nature of the characteristic function that doesn't align with our expectations from interpretations. We dwell on this in the following sections.

\textbf{Model-Specific.} By assuming complete independence within the set \bX \cite{janzing-causal} successfully separates the underlying dependencies between features from the process of interpreting a model.
This means that for any $S\subseteq P$, $\bX_S\independent \bX_{\overline{S}}$, resulting in the characteristic function $\mathbb{E}_{\bX_{\overline{S}}}\left[g(\bX_S=x_S,\bX_{\overline{S}})\right]-\bbE[g(\bX)]$ as $m$.
This is explained by constructing a bayesian network of variables $\widetilde{\bX}=\{\widetilde{X}_i:i\in[p]\}$ and $g$, where
$\widetilde{X}_i=X_i$, with the \textit{causal} direction being from right to left.

The model $g$ is used as a function of $\widetilde{\bX}$ instead of \bX, and the equivalent characteristic function is given by $\mathbb{E}\left[g(\widetilde{\bX})|do(\widetilde{\bX}_S=x_S)\right]-\bbE[g(\widetilde{\bX})]$ using Pearl's $do$ operator \cite{pearl_2009}. This transformation appears to be trivial on the outset, but bears powerful implications towards the nature of interpretations provided by the corresponding Shapley values. We analyze these implications by setting up similar examples as earlier. 
%
 \begin{small}\begin{tikzpicture}[
 node distance=0.4cm and 1.3cm,
 mynode/.style={draw,circle,text width=0.5cm,align=center}
]
\node[mynode] (gt) {$X_1$};
\node[mynode, below=of gt] (gt1) {$\widetilde{X}_1$};
\node[mynode,right=of gt] (lc) {$Y$};
\node[mynode,left=of gt] (sm) {$X_2$};
\node[mynode, below=of sm] (sm1) {$\widetilde{X}_2$};
\node[mynode,right=of lc] (cp) {$X_3$};
\node[mynode, below=of cp] (cp1) {$\widetilde{X}_3$};
\node[mynode,right=of gt1] (g) {$g$};
\path (gt) edge[-latex] (lc)
(gt) edge[-latex] (sm)
(gt) edge[-latex] (gt1)
(sm) edge[-latex] (sm1)
(lc) edge[-latex] (cp)
(cp) edge[-latex] (cp1)
(gt1) edge[-latex] (g)
(cp1) edge[-latex] (g);

\end{tikzpicture} \end{small}

Assume the variables follow $X_1\rightarrow X_2$, $X_1\rightarrow Y$, $Y\rightarrow X_3$, $X_1\rightarrow g$ and $X_3\rightarrow g$.
Based on the above construction we have\vspace{-2mm}
\begin{small}
\begin{align*}
&m(\widetilde{\bX}_{S}\cup \widetilde{X}_2) = \mathbb{E}\left[g(\widetilde{\bX})|do\left(\widetilde{\bX}_{S\cup \{2\}}=x_{S\cup\{2\}}\right)\right]-\bbE[g(\widetilde{\bX})]\\
&=\mathbb{E}\left[g(\widetilde{\bX})|do\left(\widetilde{\bX}_{S}=x_{S}\right)\right]-\bbE[g(\widetilde{\bX})]= m(\widetilde{\bX}_{S}).
\end{align*}\end{small}
for $S\subseteq \{1,3\}$. Therefore for the game $\langle \{\widetilde{X}_1,\widetilde{X}_2, \widetilde{X}_3\},m\rangle$, $\phi_2(m)=0$. Similarly if $g$ was a function of $\widetilde{X}_1$ and $\widetilde{X}_2$, $\phi_3(m)$ would be 0. This is desirable as it prevents unimportant variables with no direct effect on the model from being surfaced.

However, we must also note that this method will also result in a Shapley value of 0 for all variables which have indirect \textit{causal} effects on the model output. For example in the example with $X_1\rightarrow X_2$, $X_2\rightarrow Y$, $Y\rightarrow X_3$ and $X_2\rightarrow g$, the game $\langle \{\widetilde{X}_1,\widetilde{X}_2, \widetilde{X}_3\},m\rangle$, this method can assign a non zero score only to the player $\widetilde{X}_2$. This does not align with our expectations when we know that genes ($X_1$) do in fact affect the likelihood of a person being a smoker ($X_2$) which directly impacts the model's decision. Therefore this characteristic function fails to completely answer the question ``Why does my model believe that a person is likely to have lung cancer?''

In the previous example where  $X_1\rightarrow X_2$, $X_1\rightarrow Y$, $Y\rightarrow X_3$, $X_1\rightarrow g$ and $X_3\rightarrow g$, this method treats the variables as being independent of each other and therefore for $X_1$, it would only capture its direct influence on $g$ and fail to capture its indirect influence on $g$ via the path $X_1\rightarrow Y\rightarrow X_3\rightarrow g$.
This brings us to try and understand the finer differences between asking the following questions, \textit{(i)} ``Why did my model predict a person has lung cancer?''; \textit{(ii)} ``What caused my model to predict a person has lung cancer?'' While the former can be answered by using model-specific explanations that capture the influence of those variables which directly affect the model, as discussed in this section, the latter is discussed more in detail in the next section. 

\textbf{Causal.} To answer the question, ``What caused my model to predict a person has lung cancer?'' we are not only interested in variables which directly affect the model output, but in fact, we are also interested in the cause and effect relationships between variables. For example, when we have edges $X_1\rightarrow X_2$, $X_2\rightarrow Y$, $Y\rightarrow X_3$ and $X_2\rightarrow g$, we would also like to incorporate the indirect influence of $X_1$ on $g$ via $X_2$. $\mathbb{E}\left[g(\bX)|do(\bX_S=x_S)\right]-\bbE[g(\bX)]$ is a natural choice of characteristic function for the game $\langle \{X_1,X_2, X_3\},m\rangle$ and follows from that used for \textit{model-specific} interpretations \cite{janzing-causal}. \footnote{Our primary focus here has been on identifying the concepts governing \textit{causal} interpretations and building the theoretical machinery for programmatically computing them. We understand that \textit{causal} discovery by itself is an active area of research and here we assume that the underlying \textit{causal} Bayesian network is known.}. 
Now $\phi_3(m)$ will be 0, while $g(x_2)-\bbE[g(\bX)]$ will be distributed fairly between $\phi_1(m)$ and $\phi_2(m)$, unlike for \textit{model-specific} explanations, where $\phi_1(m)$ would also be zero. Furthermore if $g$ is also a function of $X_1$, then $\phi_1(m)$ would capture both direct as well as indirect \textit{causal} effects of $X_1$ on the model output $g$. 
 \begin{small}\begin{tikzpicture}[
 node distance=0.4cm and 0.8cm,
 mynode/.style={draw,circle,text width=0.5cm,align=center}
]
\node[mynode] (gt) {$X_1$};
\node[mynode, left=of gt] (x4) {$X_4$};
\node[mynode,right=of gt] (sm) {$X_2$};
\node[mynode,right=of sm] (lc) {$Y$};
\node[mynode,right=of lc] (cp) {$X_3$};
\node[mynode,below right=of sm] (g) {$g$};
\path (sm) edge[-latex] (lc)
(gt) edge[-latex] (sm)
(x4) edge[-latex] (gt)
(sm) edge[-latex] (g)
(lc) edge[-latex] (cp);

\end{tikzpicture} \end{small}

We now observe another variable, namely the genotype of parents ($X_4$), and a new edge corresponding to this variable is added $X_4\rightarrow X_1$. 
Here the values $\phi_i(m)$ for $i\in\{1,2\}$ will be different from earlier to ensure that $\phi_1(m)+\phi_2(m)+\phi_4(m)=m(x_1,x_2,x_3,x_4) = g(x_2)-\bbE[g(\bX)]$. $\phi_3(m)$ remains $0$.

An important point to be made here is that insights derived for the model output $g$ can be very different from those for the target variable $Y$. For example, if $g$ is a function of a variable $X_i$ which does not directly affect $Y$, then $X_i$ might surface as an important variable in the \textit{causal} or \textit{model-specific} interpretations for $g$ whereas it is irrelevant to $Y$.
%

\section{Measures of Interpretation}
\label{sec:moi}
In this section, we will discuss measures of interpretation for a learning model, of which the one used in \Sectionref{noi}, i.e. the conditional expectation of the model output, is the most widely known. Along similar lines, other measures have been proposed such as in \cite{c-shapley} with an information-theoretically sound Shapley characteristic function, and closely related \cite{janzing-cic} which defines \textit{causal} information contribution (CIC) and \textit{causal} variance contribution (CVC), for analyzing the global effects of a variable $X_i$ on the target variable $Y$. 

%
 In \Sectionref{noi}, our focus has been on carefully deciding upon the nature of interpretations we are interested in which are of three kinds, and the corresponding form of the characteristic functions accompanying it. All three forms can be applied together with different measures to obtain different types of interpretations. 
 
We will first focus on two random variables of interest, namely the model output and the predictive performance (loss function), and the application of concepts from \Sectionref{noi} to both these cases. We also discuss measures of statistical uncertainty and dispersion as informative candidates for deriving the relevance of variables towards both model output and loss.

Note that all the concepts from \Sectionref{noi} and \Sectionref{moi} are directly applicable to any random variable, for example, one could choose to examine the contribution of each variable towards the target $Y$ instead. 
 
 \subsection{Random Variable Under Consideration}
 \label{subsec:rv}
 Our focus here is on two random variables of interest, the first being the model output $g(\bX)$ which could be the scoring function of a classifier, or simply the decision in terms of the predicted classes, or it could be the predicted value by a regressor. The second random variable is the loss function $l(Y,g(\bX))$ which measures the predictive performance of $g$. 
 %
 
 \subsubsection{Model Output}
\label{subsubsec:mo}
From \Sectionref{noi}, for a data point $x$, the characteristic function for a subset of features $S\subseteq P$ in the case of \textit{observational} interpretations is given by $\bbE\big[g(\bX)|\bX_S=x_S\big]-\bbE[g(\bX)]$.
For \textit{model-specific} interpretations, we know that the characteristic function is 
$\bbE\left[g(\widetilde{\bX})|do(\widetilde{\bX}_S=x_S)\right]-\bbE[g(\bX)]$ which is equivalent to $\bbE\left[g(\bX_S=x_S, \bX_{\overline{S}})\right]-\bbE[g(\bX)]$.
And finally for \textit{causal} interpretations about the decision made by a model we have 
$\bbE\left[g(\bX)|do(\bX_S=x_S)\right]-\bbE[g(\bX)]$ which utilizes Pearl's $do$ operator \cite{pearl_2009}.

\subsubsection{Model Loss}
\label{subsubsec:ml}
The loss $l$ is a function of the model output $g$ and the target $Y$, and can be represented by an additional node in the Bayesian network presented below with incoming edges from $g$ and $Y$. Here $\widetilde{Y}=Y$ and $\widetilde{X}_i=X_i$ for $i\in P$, $P$ being $\{1,2,3\}$ in this example, and the \textit{causal} direction is from RHS to LHS. 

 \begin{small}\begin{tikzpicture}[
 node distance=0.6cm and 1.2cm,
 mynode/.style={draw,circle,text width=0.5cm,align=center},
 box/.style={draw,rectangle}
]
\node[mynode] (x1) {$X_1$};
\node[mynode,below right=of x1] (x2) {$X_2$};
\node[mynode, below =of x2] (x2t) {$\widetilde{X}_2$};
\node[mynode, below =of x1, left=of x2t] (x1t) {$\widetilde{X}_1$};
\node[mynode,above right=of x2] (y) {$Y$};
\node[mynode,right=of x2t] (yt) {$\widetilde{Y}$};
\node[mynode,right=of y] (x3) {$X_3$};
\node[mynode, below =of x3, right=of yt] (x3t) {$\widetilde{X}_3$};
\node[mynode,below=of x1t] (g) {$g$};
\node[mynode,below=of yt] (loss) {$l$};
\path (x1) edge[-latex] (x2)
(x1) edge[-latex] (x1t)
(x1) edge[-latex] (y)
(x2) edge[-latex] (y)
(x2) edge[-latex] (x2t)
(x3) edge[-latex] (x3t)
(y) edge[-latex] (x3)
(y) edge[-latex] (yt)
(x1t) edge[-latex] (g)
(x2t) edge[-latex] (g)
(g) edge[-latex] (loss)
(yt) edge[-latex] (loss);
\node[box, inner sep=2mm,fit= (x1) (y) (x2) (x3)] (b1) {};
\node[box, inner sep=2mm,fit= (x1t) (yt) (x2t) (x3t)] (b2) {};
\end{tikzpicture} \end{small}

Some well known candidates for the loss function $l$ are: a cross-entropy function when the model emits a distribution over multiple classes; the indicator function $\mathbbm{1}\{g\neq Y\}$ when $g$ is a classifier; the squared error for a regressor $g$ as follows $(g-Y)^2$. 

The game here is given by $\langle \bZ,m\rangle$ where $m$ is a characteristic function that represents the value of a subset of variables in $\bZ:=\bX\cup Y$. For the sake of notational simplicity we will represent $l$ as a function of $\bZ$ here where $\bZ=\{X_1,X_2,\ldots,X_p,Y\}$. All the concepts from \Sectionref{noi} are now directly applicable to $l$.

Referring back to \Sectionref{noi}, we can now define the characteristic functions for the \textit{observational}, \textit{model-specific} and \textit{causal} interpretations with the following characteristic functions respectively, $\bbE\left[l(\bZ)|\bZ_S=z_S\right]-\bbE[l(\bZ)]$, $\bbE\left[l(\widetilde{\bZ})|do(\widetilde{\bZ}_S=z_S)\right]-\bbE[l(\bZ)]$ and $\bbE\left[l(\bZ)|do(\bZ_S=z_S)\right]-\bbE[l(\bZ)]$.

Note that this is conceptually different from \cite{sage} in which the payoff is fairly distributed only between the set of features, by means of a characteristic function given by $\bbE\left[l(Y,\bbE\left[g(\bX)|\bX_S\right])\right]-\bbE[l(Y,\bbE[g(\bX)])]$ and is only focussed on global interpretations. This summarizes the overall \textit{observational} involvement of each feature in improving the prediction performance of $\bbE\left[g(\bX)|\bX_S\right]$ from that of a model that simply predicts the average i.e. $\bbE[g(\bX)]$. 

On the contrary, our characteristic functions naturally follow from a generic definition of interpretations for any random variable of interest, for each data point $(x,y)$. This results in a score for each variable in $\{x_1,x_2,\ldots,x_p,y\}$, signifying their contribution to $l(y,g(x))-\bbE[l(Y,g(\bX))]$ which is the surplus in terms of the deviation from the average loss. 

Interestingly, this also captures the involvement of the target ($Y$) which is particularly useful in cases where the sum of credit allocated to variables in $\bX$ is $\leq 0$, and that to $Y$ is $>0$. This signifies that the players in $\bX$ played their part in improving the model performance, however, certain unobserved factors may have influenced $Y$ in turn worsening the loss.

 \subsection{Measuring Observational Interpretations}
 \label{subsec:tom} 
 Here we will discuss other candidates for the measure of interpretation.
Shannon entropy measures the uncertainty in a random variable in terms of its unpredictability. A discrete random variable for which the symbols are unordered, naturally begets this definition and the entropy is maximized by a uniform distribution over the symbols. 
We argue that in the continuous-valued (ordered) case on restricted supports, a highly polarized distribution represents contradictory information and therefore more uncertainty as opposed to a uniform distribution which represents no information. Therefore statistical dispersion is an informative measure of the value of a coalition of features for continuous-valued random variables.

 \subsubsection{Shannon Entropy}
 \label{subsubsec:shannon_entropy}
 The Shannon entropy ($H$), which is given by $\bbE[-\log p(X)]$ where $p$ is the mass function in the discrete case, and density in the continuous case, quantifies the amount of uncertainty in a random variable in terms of its unpredictability. The less predictable the variable, the higher the uncertainty, with the uniform distribution maximizing $H$ in the discrete case as well as in the continuous case for restricted supports. 

 Assume that $Y\in\{1,2,\ldots,l\}$. And that $g$ is a multi-class classifier that predicts one of the $l$ possible values. In this case we have 
 $$m(x_S)=H(g(\bX))-H(g(\bX)|\bX_S=x_S).$$
 resulting in a fair allocation of $H(g(\bX))$ to the $p$ players. The marginal contribution of $x_i$ is given by 
 \begin{align*}
 \phi_i^S(m)&=m(x_{S\cup i})-m(x_{S})\\
 &=H(g(\bX)|\bX_S=x_S)- H(g(\bX)|\bX_{S\cup i}=x_{S\cup i}).
 \end{align*}
This can be extended to global insights using
\begin{align*}
M(S)&= \underset{x\sim\Pr(\bX)}{\mathbb{E}}[m(x_S)]\\
&=H(g(\bX))-H(g(\bX)|\bX_S)=I(\bX_S;g(\bX)).
\end{align*}
Here the notation $H(A|B):=\bbE[H(A|B=b)]$ as is conventionally used to denote conditional entropy.
The marginal contribution of $X_i$ to each $S\subseteq P-\{i\}$ is 
 \begin{align*}
 \phi_i^S(M)&=M(S\cup i)-M(S)=I(X_i;g(\bX)|\bX_S)\geq 0.
 \end{align*}
This quantity is also referred to as \textit{Causal} Information Contribution (CIC) in \cite{janzing-cic}.
Along similar lines, one would hope to use the Shannon differential entropy ($h$) for deriving insights related to the real-valued loss function. However, note that the estimation of Shannon differential entropy is non-trivial requiring us to estimate the density of a continuous random variable.
Furthermore, $h\downarrow -\infty$ for a constant random variable rendering it inapplicable as a characteristic function.

 \subsubsection{Variance}
  \label{subsubsec:variance}
 The variance of a real-valued random variable is a useful measure of dispersion and is easily computable in the continuous case. 
 For a model $g$, we can examine the model output random variable which could be the scoring function of a classifier or the prediction by a regressor, the characteristic function is given by
  $$m(x_S)=\text{Var}(g(\bX))-\text{Var}(g(\bX)|\bX_S=x_S).$$
This results in a fair distribution of $\text{Var}(g(\bX))$ between the $p$ players. Furthermore, we have 
 \begin{align*}\phi_i^S(m)=\text{Var}(g(\bX)|\bX_S=x_s)-\text{Var}(g(\bX)|\bX_{S\cup i}=x_{s\cup i}).
 \end{align*} 
 $\phi_i^S(m)$ represents the marginal change in the variance of the model output on observing $X_i=x_i$. Note that the global influence of each variable can be computed from a global characteristic function defined as 
 \begin{align*}
 M(S)&=\underset{x\sim\Pr(\bX)}{\mathbb{E}}[m(x_S)]\\
 &\hspace{-1cm}=\text{Var}(g(\bX))-\mathbb{E}\left[\text{Var}(g(\bX)|\bX_S=x_S)\right]\\
 &\hspace{-1cm}=\text{Var}\left(\mathbb{E}\left[g(\bX)|\bX_S=x_s\right]\right) \geq 0\quad\text{(by law of total variance.)}
 \end{align*} 
The marginal contribution of $X_i$ to each \begin{small}$S\subseteq P-\{i\}$\end{small} is 
 \begin{align*}
&\phi_i^S(M)= M(S\cup i)-M(S)\\
&=\bbE\left[\text{Var}\left(g(\bX)|\bX_{S}=x_{S}\right)\right]-\bbE\left[\text{Var}\left(g(\bX)|\bX_{S\cup i}=x_{S\cup i}\right)\right].
 \end{align*} 
 This quantity is also referred to as \textit{Causal} Variance Contribution (CVC) in \cite{janzing-cic}.
From the law of total variance, we know that conditioning reduces variance, as a result, $\phi_i^S(M)\geq 0$. Therefore globally, the Shapley values represent the contribution of each variable in the reduction of the variance of the model output. 
When $g(\bX)=\sum_{i=1}^{p}w_iX_i$, and $X_i$ are all independent, $\phi_i^S(M)=w_i^2\text{Var}(X_i)$.
 
 For deriving insights about the model loss, the characteristic function $m$ for the game $\langle\bZ,m\rangle$, naturally follows as
 $m(z_S)=\text{Var}(l(\bZ))-\text{Var}(l(\bZ)|\bZ_S=z_S)$.

 
\subsubsection{Cumulative Paired Shannon Entropy}
\label{subsubsec:cpse}
The cumulative paired Shannon entropy of a random variable $A$, is given by $\mathcal{E}(A)=\int_{-\infty}^{\infty}H(\mathbbm{1}\{A\leq\lambda\})d\lambda$
where $H$ is the Shannon entropy. Note that this is an easily estimable quantity using a Reimann sum.

 $\mathcal{E}$ has been unknowingly applied in several contexts such as in fuzzy set theory \cite{fuzzy_1}, uncertainty theory \cite{liu}, dispersion theory for discrete ordered data \cite{leik,yager} as well as in reliability theory \cite{rao,drissi,sunoj,navarro}. \cite{klein-1,klein-2} formalize this measure by drawing parallels to its information-theoretic counterparts, namely divergence, entropy, and mutual information. \cite{klein-1} derives the maximum entropy distributions under different constraints for $\mathcal{E}$ and formulates rank tests using the measure which coincidentally reduces to well-known rank tests from the literature as a specialization. 
 
In most cases, we encounter continuous-valued random variables such as the model output which could be the output of a regression model or the softmax output of a classifier. The model loss such as cross-entropy and mean squared error is also typically a real-valued continuous random variable. Therefore in such cases, $\mathcal{E}$ can be an informative measure of the value of a set of variables towards the model output or loss. 

\E which is a non-negative quantity possesses several properties that are analogous to that of $H$. For example, conditioning reduces \E, i.e. for two random variables $A$ and $B$, we have $\mathbb{E}[\E(A|B=b)]\leq \E(A)$. We denote $\mathbb{E}[\E(A|B=b)]$ by $\E(A|B)$ and the quantity $\E(A)-\E(A|B)$ is called cumulative paired Shannon mutual information which we denote by $\I(B\rightarrow A)$ where the arrow indicates the lack of symmetry.
The characteristic function for the model output is 
 $$m(x_S)=\mathcal{E}(g(\bX))-\mathcal{E}(g(\bX)|\bX_S=x_S).$$ 
resulting in a fair allocation of $\E(g(\bX))$ between $p$ players.
The marginal contribution of $x_i$ is given by 
 \begin{align*}
 \phi_i^S(m)&=m(x_{S\cup i})-m(x_s)\\
 &=\E(g(\bX)|\bX_S=x_S)- \E(g(\bX)|\bX_{S\cup i}=x_{S\cup i}).
 \end{align*} 
For deriving global insights we use the following characteristic function
 \begin{align*}
M(S)&= \underset{x\sim\Pr(\bX)}{\mathbb{E}}[m(x_S)]=\E(g(\bX))-\E(g(\bX)|\bX_S)\\
&=\I(\bX_S\rightarrow g(\bX)).
\end{align*} 
The marginal contribution of $X_i$ to each \begin{small}$S\subseteq P-\{i\}$\end{small} is
 \begin{align*}
 \phi_i^S(M)&=M(S\cup i)-M(S)=\I(X_i\rightarrow g(\bX)|\bX_S)\geq 0.
 \end{align*} 
We call this Cumulative Paired Information Contribution (CPIC).

 For deriving insights pertaining to the model loss, the characteristic function $m$ for the game $\langle\bZ,m\rangle$, is given by
 $m(z_S)=\E(l(\bZ))-\E(l(\bZ)|\bZ_S=z_S)$.
  
 \subsection{Measuring Model-Specific Interpretations}
 \label{subsec:mmsi} 
%
For measures in terms of the Shannon entropy, variance, and cumulative paired Shannon entropy, for the model output we use $m(x_S)=\xi(g(\widetilde{\bX}))-\xi(g(\widetilde{\bX})|do(\widetilde{\bX}_S=x_S))$ where $\xi\in\{\text{Var}, H, \E\}$ respectively. This is equivalent to
$m(x_S)=\xi(g(\bX))-\xi(g(\bX_S=x_S,\bX_{\overline{S}})))$.

Along similar lines by replacing $g,\bX,x$ with $l,\bZ,z$ defined in \Subsubsectionref{ml}, in the characteristic function and computing Shapley values for the new game $\langle\bZ,m\rangle$, we can derive \textit{model-specific} interpretations for the loss.

 \subsection{Measuring Causal Interpretations}
 \label{subsec:causal_int} 
 Following from \Subsectionref{mmsi}, the characteristic function used for deriving insights about the model output is given by $m(x_S)=\xi(g(\bX))-\xi(g(\bX)|do(\bX_S=x_S))$ where $\xi\in\{\text{Var}, H, \E\}$. The game for deriving \textit{causal} interpretations about the loss function is given by $\langle\bZ,m\rangle$ where $m(z_S)=\xi(l(\bZ))-\xi(l(\bZ)|do(\bZ_S=z_S))$ and $\bZ$ is as defined in \Subsubsectionref{ml}.

 \subsection{Note on comparison with SAGE \cite{sage}}
 \label{subsec:sage_comp}
 In \cite[Section C.1]{sage}, we see that for the optimal Bayes classifier, the SAGE importance score is given by computing Shapley values for the game $\langle \bX,M\rangle$, where $M(S)=I(X_S;Y)$. The marginal contribution of $X_i$ globally is $M(S\cup i)-M(S)=I(X_i;Y|\bX_S)$.
Using the measure from \Subsubsectionref{shannon_entropy} for the target $Y$ as the random variable of interest, we arrive at the same characteristic function $M(S)=I(X_S;Y)$.
The equivalence of SAGE for optimal regression models \cite[Section C.2]{sage} with \Subsubsectionref{variance} applied for $Y$, also holds.

Therefore the SAGE importance scores for optimal models summarize the predictive contribution of each feature in reducing the uncertainty in the target variable $Y$. For a generic model, SAGE summarizes feature relevance in improving the predictive performance of the model $g$ from that of the performance of the average model output $\bbE[g(\bX)]$ (see \Appendixref{app2} for more details). 
On the other hand, our characteristic functions about loss explain the contribution of each element in $\{X_1,\ldots,X_p,Y\}$ towards the discrepancy in the prediction performance for a particular data point from the average prediction performance, which, as one would expect, could also be attributed to $Y$.

\section{Experiments}
\label{sec:results}
We evaluate the benefits of our interpretations for the model output, by performing standard tasks such as supervised clustering as done in \cite{lundberg} and assessing model sensitivity to different features such as in \cite{c-shapley,lundberg_shap}\footnote{We emphasize that all measures of interpretation from \Sectionref{moi} are theoretically significant in their own right, with some outperforming others in select tasks.}. \Appendixref{num}  demonstrates the differences between \textit{observational}, \textit{model-specific} and \textit{causal} interpretations using a simple example. Here we present a comparison between different measures from \Sectionref{moi} for \textit{model-specific} interpretations
\footnote{The other two variations from \Sectionref{noi}, present challenges in empirical estimation, where \textit{observational} interpretations requires the conditional density and \textit{causal} interpretations requires the underlying \textit{causal} graph.}.

We employed the weighted linear regression formulation from \cite{charnes1988}, as also done in \cite{lundberg_shap,c-shapley,janzing-causal}, for exactly computing Shapley values. Similar to KernelSHAP \cite{lundberg_shap}, the solution is approximated by solving a linear regression objective with terms sampled from the original weighted linear regression, according to the distribution of the weights. For a fair comparison, we maintain the same number of sampled terms and a sufficiently large background data set across different measures of interpretation, which amounts to a constant number of model evaluations for each data point. 

Comparisons are made between different measures from \Sectionref{moi} not only towards the model output but also towards the model loss. This automatically covers KernelSHAP \cite{lundberg_shap} which explains the model output $g$ in terms of the expected value, and in addition, we also compare against LIME \cite{lime}. We performed our experiments on a $k$-nearest neighbors classifier trained on the Census Income data set \cite{uci}, a random forest regressor trained on the Boston housing prices data set \cite{uci}, an AdaBoost classifier trained on the breast cancer data set \cite{uci} and an SVM classifier on the handwritten digits data set \cite{uci}. Here we present a select few results and provide more details in \Appendixref{app1}.\vspace{-2mm}
\begin{figure}[h!]
\centering
 \includegraphics[width=0.7\linewidth]{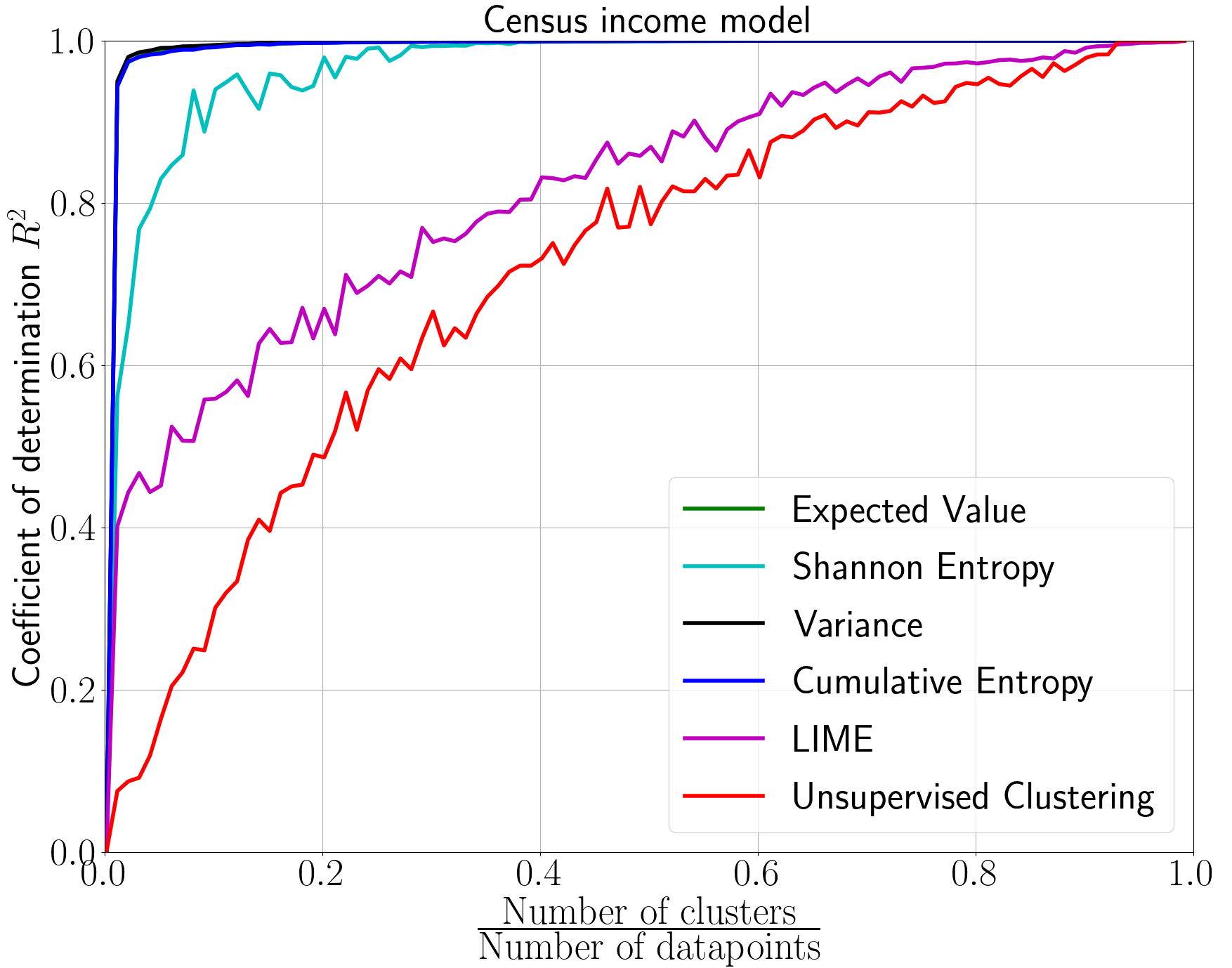}
\caption{\textbf{Supervised clustering performance} Interpretations generated for the $k$-NN classifier output are clustered. Expected value corresponds to KernelSHAP\cite{lundberg_shap}. The other measures are based on Shannon entropy $H$ \Subsubsectionref{shannon_entropy}, variance \Subsubsectionref{variance}, and cumulative paired Shannon entropy \E \Subsubsectionref{cpse}. Unlike for unsupervised clustering and LIME, all other forms of interpretation achieve a higher $R^2$ with very few clusters.}
\label{fig:supervised_census_income_model}
\end{figure}

\textbf{Supervised Clustering:} $k$-means is a well-known algorithm for unsupervised clustering. For evaluating a clustering, we can use the average prediction of all points in a cluster as the prediction for each point in the cluster. Its quality is determined by the fraction of the variance in the model output ($Y$) that is explained by the clustering, given by the coefficient of determination $R^2$. 
The underlying premise of supervised clustering is that two points with similar interpretations will have a similar model response. Therefore, a good supervised clustering algorithm will require fewer clusters for achieving a high $R^2$. \Figureref{supervised_census_income_model} presents a quantitative evaluation of different measures of interpretation based on their supervised clustering performance where we find that Shapley symmetrization on measures from \Sectionref{moi}, including KernelSHAP, clearly outperforms LIME and unsupervised clustering.\vspace{-2mm}
\begin{figure}[h!]
\centering
 \includegraphics[width=0.7\linewidth]{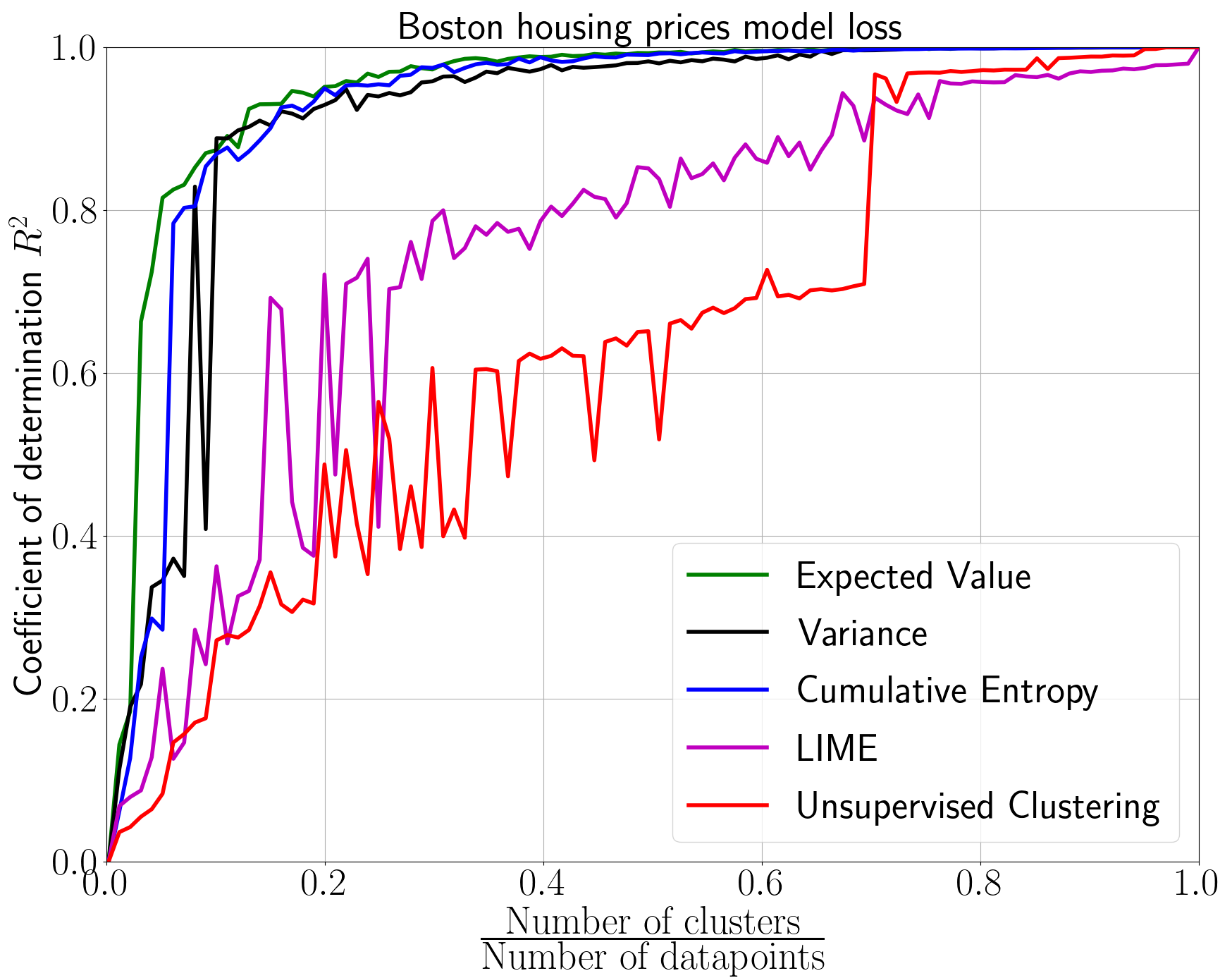}
\caption{\textbf{Supervised clustering for model loss} Here the $R^2$ corresponds to the percentage of the variance of the mean squared error loss ($l$) that is explained by the clustering of $p+1$ dimensional interpretations. Measures based on expected value (KernelSHAP), cumulative entropy, and variance produce superior clusters. 
}
\label{fig:supervised_boston_housing_loss}
\end{figure}

We also perform supervised clustering for evaluating interpretations derived from characteristic functions about the loss function \Subsubsectionref{ml} in \Figureref{supervised_boston_housing_loss}. The clustering is performed on the interpretations which are $p+1$ dimensional and the $R^2$ measures the fraction of explained variance in the loss ($l$). 
We find that our interpretations require only a few clusters to explain a large fraction of the variance in $l$. In these examples, LIME requires a significantly larger number of clusters for achieving the same $R^2$, and unsupervised clustering is the least efficient as expected.

\begin{figure}[h!]
\centering
 \includegraphics[width=0.7\linewidth]{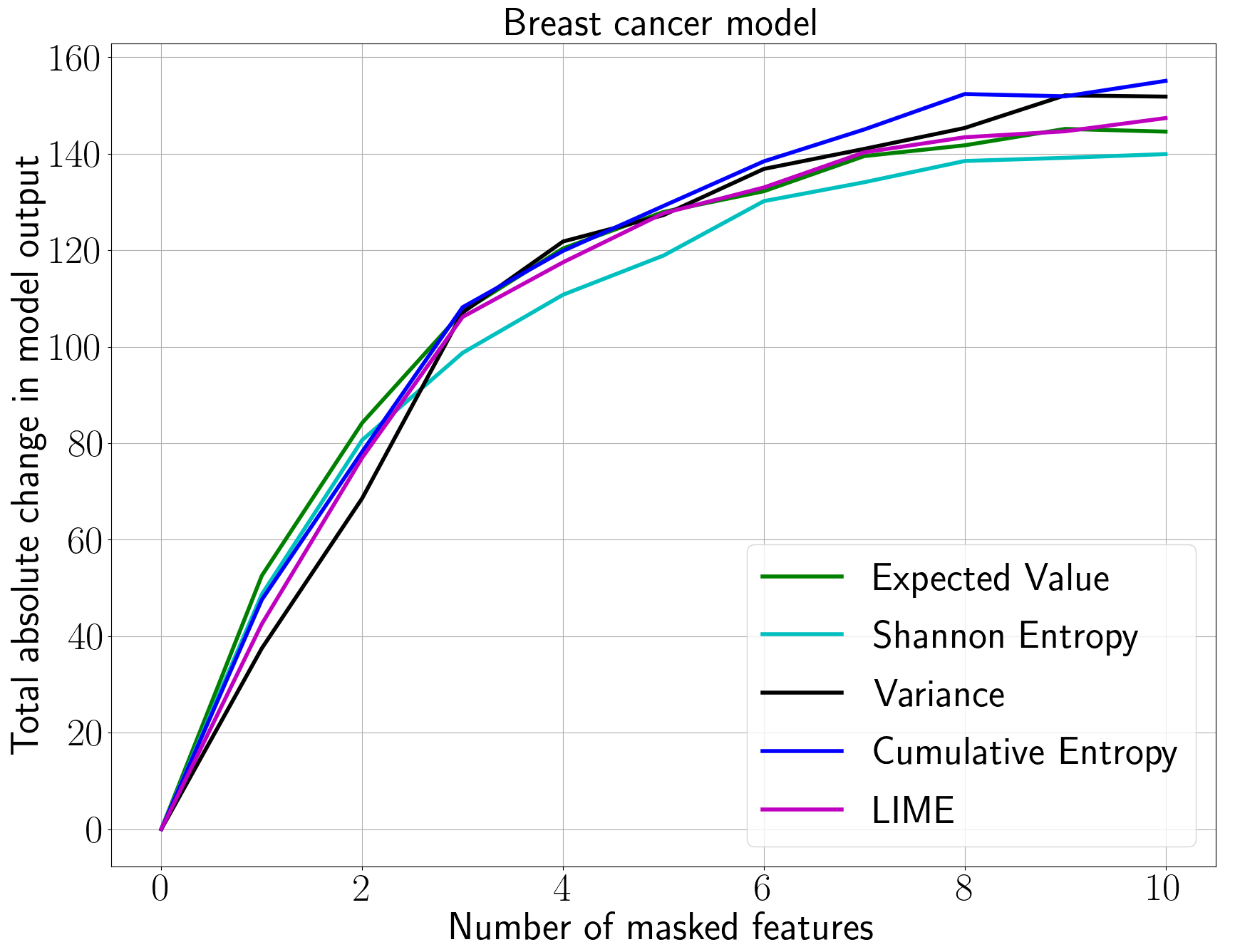}
\caption{\textbf{Model Sensitivity} We examine the sensitivity of the model to masking important features as determined by different methods. We find that Shapley values derived using different characteristic functions can identify important features with the superior ones being those based on cumulative entropy, variance, and expected value (KernelSHAP) in this example. This is not always the case as we see in \Figureref{digits}. 
}
\label{fig:sensitivity_breast_cancer_model}
\end{figure}

\textbf{Model Sensitivity:} We mask the model inputs successively based on the feature attributions given by the different methods of interpretation and study how much the model output changes. In \Figureref{sensitivity_breast_cancer_model} we see that measures based on the expected value (KernelSHAP), Shannon entropy ($H$), and cumulative paired Shannon entropy ($\E$) identify the top two features. By the tenth feature, we find $\E$, and the variance-based measure outperforming the others by a modest margin. \E does indeed also perform well for the SVM classifier trained on the digits data set in \Figureref{digits} (\Appendixref{app1}). 
LIME exhibits comparable performance to other methods, which implies that it can effectively order features by importance even though the values are not locally accurate. 

\section{Conclusion}
\label{sec:conclusion}
Providing a clear description of what interpretability entails drives the discussion in this paper. We separately present two components of interpretability: nature of interpretation \Sectionref{noi}, and measure of interpretation \Sectionref{moi} which fit together into the Shapley solution concept and provide meaningful insights. We present a generic framework for attributing relevance to different variables in interpreting the model output and predictive performance. Our numerics approximate Shapley values, however, the computational burden can potentially be alleviated under certain Markovian assumptions as done in \cite{c-shapley} for structured data.

\begin{acknowledgments}
~The author is grateful to Deepak Pai for helpful discussions and feedback amounting to significant improvements in the early stages of this work.
\end{acknowledgments}

\bibliographystyle{IEEEtran}
\bibliography{reference}
\newpage
\appendix

\section{Numerical Example}
\label{app:num}
\begin{figure*}[h!]
\centering
\begin{tikzpicture}[
  node distance=2cm and 2cm,
  mynode/.style={draw,ellipse,text width=2cm,align=center}
]
\node[mynode] (sm) {Smoker};
\node[mynode,below =of sm] (ca) {Cancer};
\node[mynode,below =of ca] (dy) {Dyspnoea};
\node[mynode,left=of dy] (xr) {X Ray};
\node[mynode,right=of ca] (g) {Model};
\node[mynode,right=of dy] (l) {Loss};
\path (sm) edge[-latex] (ca)
(ca) edge[-latex] (xr) 
(ca) edge[-latex] (dy)
(sm) edge[-latex] (g) 
(ca) edge[-latex] (g) 
(dy) edge[-latex] (l) 
(g) edge[-latex] (l) ;
\node[left=0.5cm of sm]
{
\begin{tabular}{M{1}M{1}}
\toprule
\multicolumn{2}{c}{Smoker} \\
\multicolumn{1}{c}{1} & \multicolumn{1}{c}{0} \\
\cmidrule{1-2}
0.3 & 0.7 \\
\bottomrule
\end{tabular}
};

\node[left=0.5cm of ca]
{
\begin{tabular}{cM{2}M{2}}
\toprule
& \multicolumn{2}{c}{Cancer} \\
Smoker & \multicolumn{1}{c}{1} & \multicolumn{1}{c}{0} \\
\cmidrule(r){1-1}\cmidrule(l){2-3}
0 & 0.003 & 0.997 \\
1 & 0.032 & 0.968 \\
\bottomrule
\end{tabular}
};

\node[above=0.8cm of g]
{
\begin{tabular}{ccM{2}M{2}}
\toprule
& & \multicolumn{2}{c}{Model} \\
\multicolumn{2}{l}{Smoker Cancer} & \multicolumn{1}{c}{1} & \multicolumn{1}{c}{0} \\
\cmidrule(r){1-2}\cmidrule(l){3-4}
0 & 0 & 0 & 1 \\
0 & 1 & 0 & 1 \\
1 & 0 & 0 & 1 \\
1 & 1 & 1 & 0 \\
\bottomrule
\end{tabular}
};

\node[below=0.5cm of xr]
{
\begin{tabular}{cM{2}M{2}}
\toprule
& \multicolumn{2}{c}{X Ray} \\
Cancer & \multicolumn{1}{c}{1} & \multicolumn{1}{c}{0} \\
\cmidrule(r){1-1}\cmidrule(l){2-3}
0 & 0.2 & 0.8 \\
1 & 0.9 & 0.1 \\
\bottomrule
\end{tabular}
};

\node[below=0.5cm of dy]
{
\begin{tabular}{cM{2}M{2}}
\toprule
& \multicolumn{2}{c}{Dyspnoea} \\
Cancer & \multicolumn{1}{c}{1} & \multicolumn{1}{c}{0} \\
\cmidrule(r){1-1}\cmidrule(l){2-3}
0 & 0.3 & 0.7 \\
1 & 0.65 & 0.35 \\
\bottomrule
\end{tabular}
};

\node[below=0.5cm of l]
{
\begin{tabular}{ccM{2}M{2}}
\toprule
& & \multicolumn{2}{c}{Loss} \\
\multicolumn{2}{l}{Dyspnoea Model} & \multicolumn{1}{c}{1} & \multicolumn{1}{c}{0} \\
\cmidrule(r){1-2}\cmidrule(l){3-4}
0 & 0 & 0 & 1 \\
0 & 1 & 1 & 0 \\
1 & 0 & 1 & 0 \\
1 & 1 & 0 & 1 \\
\bottomrule
\end{tabular}
};

\end{tikzpicture}
\caption{Causal Bayesian Network representing the relationships between variables related to Smoking, along with a trained model for predicting the symptom. Dyspnoea and the associated classification error as loss.}
\label{fig:bn_smoking}
\end{figure*}
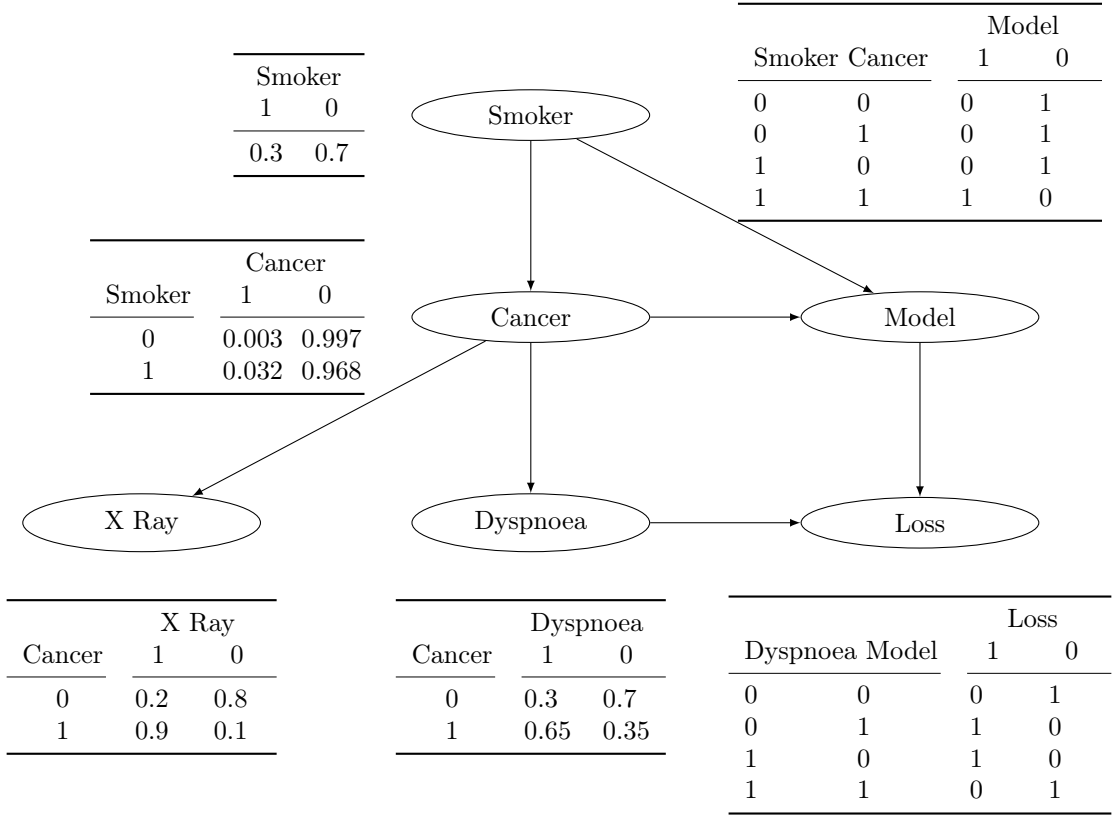
In this section, we will illustrate the differences between \textit{observational}, \textit{model-specific} and \textit{causal} explanations described in \Sectionref{noi} with a simple example. 

The \textit{causal} Bayesian network in \Figureref{bn_smoking} captures the directional dependencies between Smoker ($X_1$), Cancer ($X_2$), X Ray ($X_3$) and Dyspnoea ($Y$) all of which are indicator random variables. 

In addition it also has a model ($g$) which is a predictor for Dyspnoea and predicts 1 only if the person is a smoker and has cancer as described by the corresponding conditional probability table. The loss ($l$) is an indicator for when the model misclassifies the person as exhibiting Dyspnoea or not. 

Now consider the data point $z=[x_1,x_2,x_3,y]=[1,1,1,0]$. We will use the expected value as given in \Subsectionref{rv} as the measure of interpretation for this particular data point.

\subsection{Model Output}
We will first focus on different types of interpretations sought for the model output $g$. We are interested in the game $\langle\bX,m\rangle$ where $m$ is the characteristic function that is completely defined by the nature of interpretation \Sectionref{noi} sought and the measure of interpretation of significance \Sectionref{moi}.

For computing the \textit{observational} interpretations we must first evaluate the characteristic function for each coalition given by $m(x_S)=\bbE\left[g(\bX)|\bX_S=x_S\right]-\bbE[g(\bX)]$. 
$m(x_\emptyset)=0,
m(x_{1})=0.0224,
m(x_{2})=0.8109,
m(x_{3})=0.0319,
m(x_{1,2})=0.9904,
m(x_{2,3})=0.8109,
m(x_{1,3})=0.1199,
m(x_{1,2,3})=0.9904$.
For answering the question ``How are my variables associated with the predictor for Dyspnoea?'' we compute \textit{observational} interpretations. $$[\phi_1(m),\phi_2(m), \phi_3(m)]=[0.1119,0.8516,0.0269]$$
Here we can see that the person having cancer is most associated with the positive model output. This is because its association with the model output is felt not only directly but also via Smoker. X Ray is the least associated with the model predicting that the person has Dyspnoea, however the attribution is non zero which was pointed out as a problem in \cite{ig} because X Ray does not affect the model $g$. 

To answer this discrepancy, we need to instead ask the following question, ``Why did the model for Dypnoea predict positive''. This is done by computing \textit{model-specific} interpretations for which we first evaluate $m(x_S)=\bbE\left[g(\bX_S=x_S, \bX_{\overline{S}})\right]-\bbE[g(\bX)]$.
$m(x_\emptyset)=0,
m(x_{1})=0.0021,
m(x_{2})=0.2904,
m(x_{3})=0,
m(x_{1,2})=0.9904,
m(x_{2,3})=0.2904,
m(x_{1,3})=0.0021,
m(x_{1,2,3})=0.9906$.
%
On computing \textit{model specific} interpretations, we get $$[\phi_1(m),\phi_2(m), \phi_3(m)]=[0.3510,0.6394,0]$$ 
This captures the direct influence of each variable on the model output. Since X Ray does not affect the predictor, it gets a score of 0. In addition, the disparity between Smoker and Cancer is reduced since only the direct influence of each of them is being considered. 

Note that in the above case, we are missing the influence that the variable Smoker had on the model output through its influence on Cancer. In our example, the Smoking may have caused cancer which in turn is an important cause for the model to predict that the person has Dyspnoea. The question to be answered here is ``What is the involvement of each variable in causing my model to predict that the person has Dyspnoea?''. We compute \textit{causal} interpretations for we must first evaluate $m(x_S)=\bbE\left[g(\bX)|do(\bX_S=x_S)\right]-\bbE[g(\bX)]$.
$m(x_\emptyset)=0,
m(x_{1})=0.0224,
m(x_{2})=0.2904,
m(x_{3})=0,
m(x_{1,2})=0.9904,
m(x_{2,3})=0.2904,
m(x_{1,3})=0.0224,
m(x_{1,2,3})=0.9906$.
On computing \textit{causal} interpretations, we get $$[\phi_1(m),\phi_2(m), \phi_3(m)]=[0.3612,0.6292,0]$$ 
Observe that the attribution for Smoker has increased and that for Cancer decreased further. This is because more of the surplus than earlier, can be attributed to Smoker which was also involved in causing Cancer thereby affecting the model indirectly.

\subsection{Model Loss}
Here we will focus on different types of interpretations sought for the model loss $l=\mathbbm{1}\{g\neq Y\}$. We are interested in the game $\langle\bZ,m\rangle$ where $\bZ:=\{Z_1,Z_2,Z_3,Z_4\}=\{X_1,X_2,X_3,Y\}$.
For computing the \textit{observational} interpretations we will first evaluate the characteristic function $m(z_S)=\bbE\left[l(\bZ)|\bZ_S=z_S\right]-\bbE[l(\bZ)]$.
$m(z_\emptyset)=0,
m(z_{1})=0.0004,
m(z_{2})=0.1026,
m(z_{3})=0.0041,
m(z_{4})=-0.2964,
m(z_{1,2})=0.0488,
m(z_{2,3})=0.1026,
m(z_{1,3})=0.0053,
m(z_{1,4})=-0.2849,
m(z_{2,4})=0.5193,
m(z_{3,4})=-0.2799,
m(z_{1,2,3})=0.0488,
m(z_{1,2,4})=0.6988,
m(z_{2,3,4})=0.5193,
m(z_{1,3,4})=-0.2322,
m(z_{1,2,3,4})=0.6988$.
On computing \textit{observational} interpretations, we get \begin{small}$$[\phi_1(m),\phi_2(m), \phi_3(m),\phi_4(m)]=[0.056,0.4908,0.0072,0.1448]$$\end{small}
Here we observe that X Ray has a positive associative effect, given by 0.0072, on the loss and the target Dyspnoea also has a positive associative effect on the loss. The importance of Smoker is small compared to the other variables.

For computing \textit{model-specific} interpretations we first evaluate the characteristic function $m(z_S)=\bbE\left[l(\widetilde{\bZ})|do(\widetilde{\bZ}_S=z_S)\right]-\bbE[l(\bZ)]$.
$m(z_\emptyset)=0,
m(z_{1})=-0.0006,
m(z_{2})=0.1162,
m(z_{3})=0,
m(z_{4})=-0.2916,
m(z_{1,2})=0.3947,
m(z_{2,3})=0.1162,
m(z_{1,3})=-0.0006,
m(z_{1,4})=-0.2895,
m(z_{2,4})=-0.0012,
m(z_{3,4})=-0.2916,
m(z_{1,2,3})=0.6959-0.3012,
m(z_{1,2,4})=0.6988,
m(z_{2,3,4})=-0.0012,
m(z_{1,3,4})=-0.2895,
m(z_{1,2,3,4})=0.6988$.
On computing \textit{model-specific} interpretations, we get \begin{small}$$[\phi_1(m),\phi_2(m), \phi_3(m),\phi_4(m)]=[0.2799,0.4824,0,-0.0635]$$\end{small}
Here we see that the importance of Smoker is magnified which is in fact its true direct effect on the loss. As is expected X Ray has no direct effect on the loss. And the target Dyspnoea seems to be playing a small part in improving the loss for this data point.

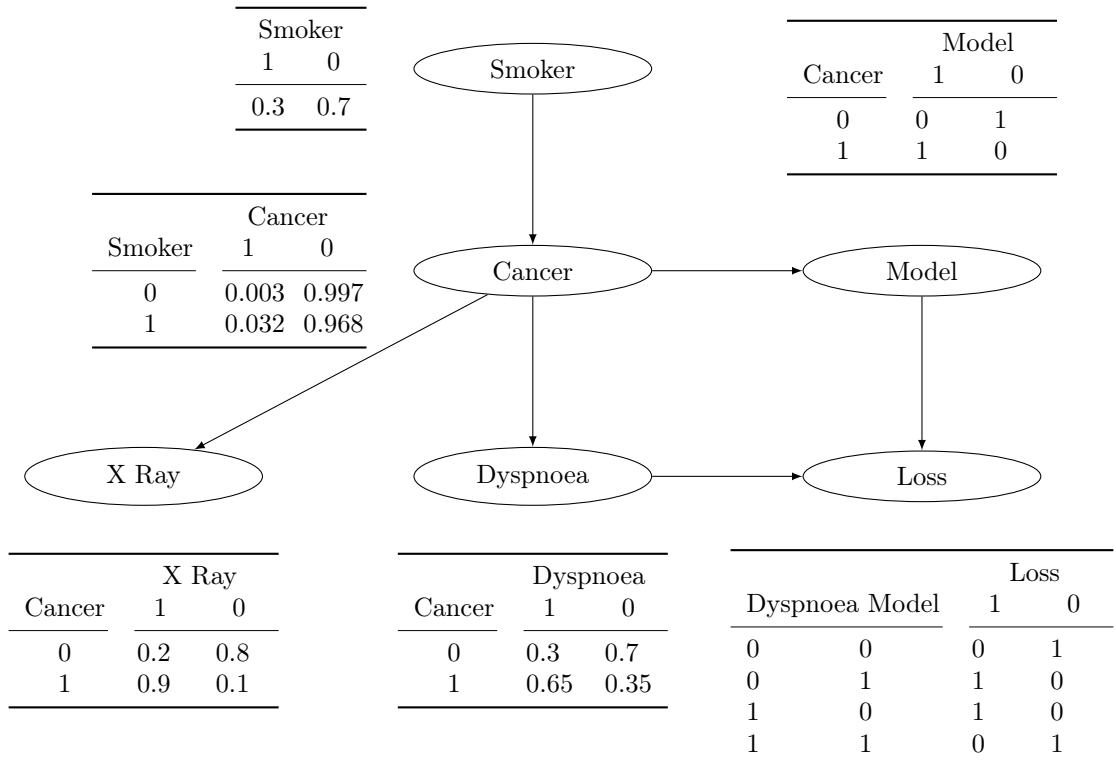
\begin{figure*}[h!]
\centering
\begin{tikzpicture}[
  node distance=2cm and 2cm,
  mynode/.style={draw,ellipse,text width=2cm,align=center}
]
\node[mynode] (sm) {Smoker};
\node[mynode,below =of sm] (ca) {Cancer};
\node[mynode,below =of ca] (dy) {Dyspnoea};
\node[mynode,left=of dy] (xr) {X Ray};
\node[mynode,right=of ca] (g) {Model};
\node[mynode,right=of dy] (l) {Loss};
\path (sm) edge[-latex] (ca)
(ca) edge[-latex] (xr) 
(ca) edge[-latex] (dy)
(ca) edge[-latex] (g) 
(dy) edge[-latex] (l) 
(g) edge[-latex] (l) ;
\node[left=0.5cm of sm]
{
\begin{tabular}{M{1}M{1}}
\toprule
\multicolumn{2}{c}{Smoker} \\
\multicolumn{1}{c}{1} & \multicolumn{1}{c}{0} \\
\cmidrule{1-2}
0.3 & 0.7 \\
\bottomrule
\end{tabular}
};

\node[left=0.5cm of ca]
{
\begin{tabular}{cM{2}M{2}}
\toprule
& \multicolumn{2}{c}{Cancer} \\
Smoker & \multicolumn{1}{c}{1} & \multicolumn{1}{c}{0} \\
\cmidrule(r){1-1}\cmidrule(l){2-3}
0 & 0.003 & 0.997 \\
1 & 0.032 & 0.968 \\
\bottomrule
\end{tabular}
};

\node[above=0.8cm of g]
{
\begin{tabular}{cM{2}M{2}}
\toprule
& \multicolumn{2}{c}{Model} \\
Cancer & \multicolumn{1}{c}{1} & \multicolumn{1}{c}{0} \\
\cmidrule(r){1-1}\cmidrule(l){2-3}
0 & 0 & 1 \\
1 & 1 & 0 \\
\bottomrule
\end{tabular}
};


\node[below=0.5cm of xr]
{
\begin{tabular}{cM{2}M{2}}
\toprule
& \multicolumn{2}{c}{X Ray} \\
Cancer & \multicolumn{1}{c}{1} & \multicolumn{1}{c}{0} \\
\cmidrule(r){1-1}\cmidrule(l){2-3}
0 & 0.2 & 0.8 \\
1 & 0.9 & 0.1 \\
\bottomrule
\end{tabular}
};

\node[below=0.5cm of dy]
{
\begin{tabular}{cM{2}M{2}}
\toprule
& \multicolumn{2}{c}{Dyspnoea} \\
Cancer & \multicolumn{1}{c}{1} & \multicolumn{1}{c}{0} \\
\cmidrule(r){1-1}\cmidrule(l){2-3}
0 & 0.3 & 0.7 \\
1 & 0.65 & 0.35 \\
\bottomrule
\end{tabular}
};

\node[below=0.5cm of l]
{
\begin{tabular}{ccM{2}M{2}}
\toprule
& & \multicolumn{2}{c}{Loss} \\
\multicolumn{2}{l}{Dyspnoea Model} & \multicolumn{1}{c}{1} & \multicolumn{1}{c}{0} \\
\cmidrule(r){1-2}\cmidrule(l){3-4}
0 & 0 & 0 & 1 \\
0 & 1 & 1 & 0 \\
1 & 0 & 1 & 0 \\
1 & 1 & 0 & 1 \\
\bottomrule
\end{tabular}
};

\end{tikzpicture}
\caption{Causal Bayesian Network analogous to \Figureref{bn_smoking} but with a different predictor for Dyspnoea denoted by Model.}
\label{fig:bn_smoking_1}
\end{figure*}

For computing \textit{causal} interpretations we first evaluate $m(z_S)=\bbE\left[l(\bZ)|do(\bZ_S=z_S)\right]-\bbE[l(\bZ)]$.
$m(z_\emptyset)=0,
m(z_{1})=-0.0006,
m(z_{2})=0.2588,
m(z_{3})=0,
m(z_{4})=-0.2916,
m(z_{1,2})=0.0488,
m(z_{2,3})=0.2588,
m(z_{1,3})=-0.0006,
m(z_{1,4})=-0.2692,
m(z_{2,4})=-0.0012,
m(z_{3,4})=-0.2916,
m(z_{1,2,3})=0.0488,
m(z_{1,2,4})=0.6988,
m(z_{2,3,4})=-0.0012,
m(z_{1,3,4})=-0.2692,
m(z_{1,2,3,4})=0.6988$.
%
On computing \textit{causal} interpretations, we get \begin{small}$$[\phi_1(m),\phi_2(m), \phi_3(m),\phi_4(m)]=[0.2018,0.4656,0,0314]$$\end{small}
From here our takeaway is that the \textit{causal} contribution towards the loss for $x$ being worse than average, can be attributed only to the variables smoker, cancer and Dyspnoea. Furthermore, unlike what the \textit{observational} scores might suggest, the variable smoker does play a much more significant part in worsening the loss, however not as much as the analysis based only on direct effects (\textit{model-specific}) might suggest. The target variable, Dyspnoea itself had a small worsening effect on the loss. 
\subsection{Interpretations for a Different Model (\Figureref{bn_smoking_1})}
In the previous sections the model was given by $\text{Model} = \text{Smoker} \& \text{Cancer}$, with the causal direction being right to left. We now consider another model, given by $\text{Model} = \text{Cancer}$, with the causal direction being from right to left again. 
Here the interpretations towards the model output are given by \\
1) \textit{Observational}: $[\phi_1(m),\phi_2(m), \phi_3(m)]=[0.0199,0.9392,0.0293]$\\
2) \textit{Model-Specific}: $[\phi_1(m),\phi_2(m), \phi_3(m)]=[0,0.9884,0]$\\
3) \textit{Causal}: $[\phi_1(m),\phi_2(m), \phi_3(m)]=[0.0102,0.9782,0]$

As the \textit{observational} interpretations suggest, Cancer is the most relevant variable towards the model output. Furthermore, we see that the \textit{model-specific} interpretations for Smoker and X Ray are 0, and this is because both these variables do not have any direct effects on the model output. What would also be interesting is to derive the \textit{causal} effect of Smoker on the model output which follows an indirect path via Cancer. The derived \textit{causal} interpretations quantify this as 0.0102 for Smoker. Once again, as expected, X Ray does not surface as important.

As done for the model output, we also derive the interpretations for the loss. \\
1) \textit{Observational}: $[\phi_1(m),\phi_2(m), \phi_3(m),\phi_4(m)]=[0.0051,0.5027,0.0069,0.1847]$\\
2) \textit{Model-Specific}: $[\phi_1(m),\phi_2(m), \phi_3(m),\phi_4(m)]=[0,0.6918,0,0.0076]$\\
3) \textit{Causal}: $[\phi_1(m),\phi_2(m), \phi_3(m),\phi_4(m)]=[0.0037,0.6849,0,0.0108]$

Quite unsurprisingly, we see that X Ray is deemed irrelevant by both \textit{model-specific} and \textit{causal} interpretations as it has no direct or indirect effects on the loss. Cancer is attributed as the most significant variable towards the misclassification. Smoker has a small involvement towards the misclassification as compared to Cancer, as given by the \textit{causal} attribution 0.0037. We find that the overall relevance of Smoker is lower for \Figureref{bn_smoking_1} than for \Figureref{bn_smoking}. This is because the direct effects of Smoker on the model output are more profound than the indirect effects, for the model given by $\text{Model} = \text{Smoker} \& \text{Cancer}$. In \Figureref{bn_smoking_1}, the variable, Smoker affects the model only indirectly via the variable Cancer.

\newpage
\section{Note on SAGE \cite{sage} (continued from \Subsectionref{sage_comp})}
\label{app:app2}
\Subsubsectionref{ml} and \Subsectionref{sage_comp} highlight the main distinctions between the interpretations about loss generated by our characteristic functions and those generated by Shapley Additive Global Importance (SAGE \cite{sage}). 

To reiterate, SAGE summarizes relevance of each feature i.e. $\{X_1,\ldots,X_p\}$, in improving the predictive performance of the model $g$ from that of the performance of the average model output $\bbE[g(\bX)]$, i.e. the contributions to $l(y,g(x))-l(y,\bbE[g(\bX)])$.

On the other hand, our characteristic functions about loss explain the contribution of each element in $\{X_1,\ldots,X_p,Y\}$ towards the discrepancy in the prediction performance for a particular data point from the average prediction performance, i.e. the contributions to $l(y,g(x))-\bbE[l(Y,g(\bX))]$.

\Subsubsectionref{ml} discusses different characteristic functions for interpreting the relevance of each variable to the loss for each data point, in terms of the latter. Similarly here we discuss how \textit{observational}, \textit{model-specific} and \textit{causal} interpretations (see \Sectionref{noi}) are generated for a local variant of SAGE, i.e. for each data point. 

The characteristic function for the \textit{observational} interpretations is given by $l(y,\bbE[g(\bX)|\bX_S=x_s])-l(y,\bbE[g(\bX)])$. For deriving \textit{model-specific} interpretations, the characteristic function to be used is $l(y,\bbE[g(\widetilde{\bX})|do(\widetilde{\bX}_S=x_s)])-l(y,\bbE[g(\bX)])$, and, that for \textit{causal} interpretations is given by $l(y,\bbE[g(\bX)|do(\bX_S=x_s)])-l(y,\bbE[g(\bX)])$.

Note that both the above variants of interpretations about the loss for each data point have significantly different implications with no superior between the two.

\newpage
\section{Omitted plots from \Sectionref{results}}
\label{app:app1}
\begin{figure}[h!]
\centering
 \includegraphics[width=0.7\linewidth]{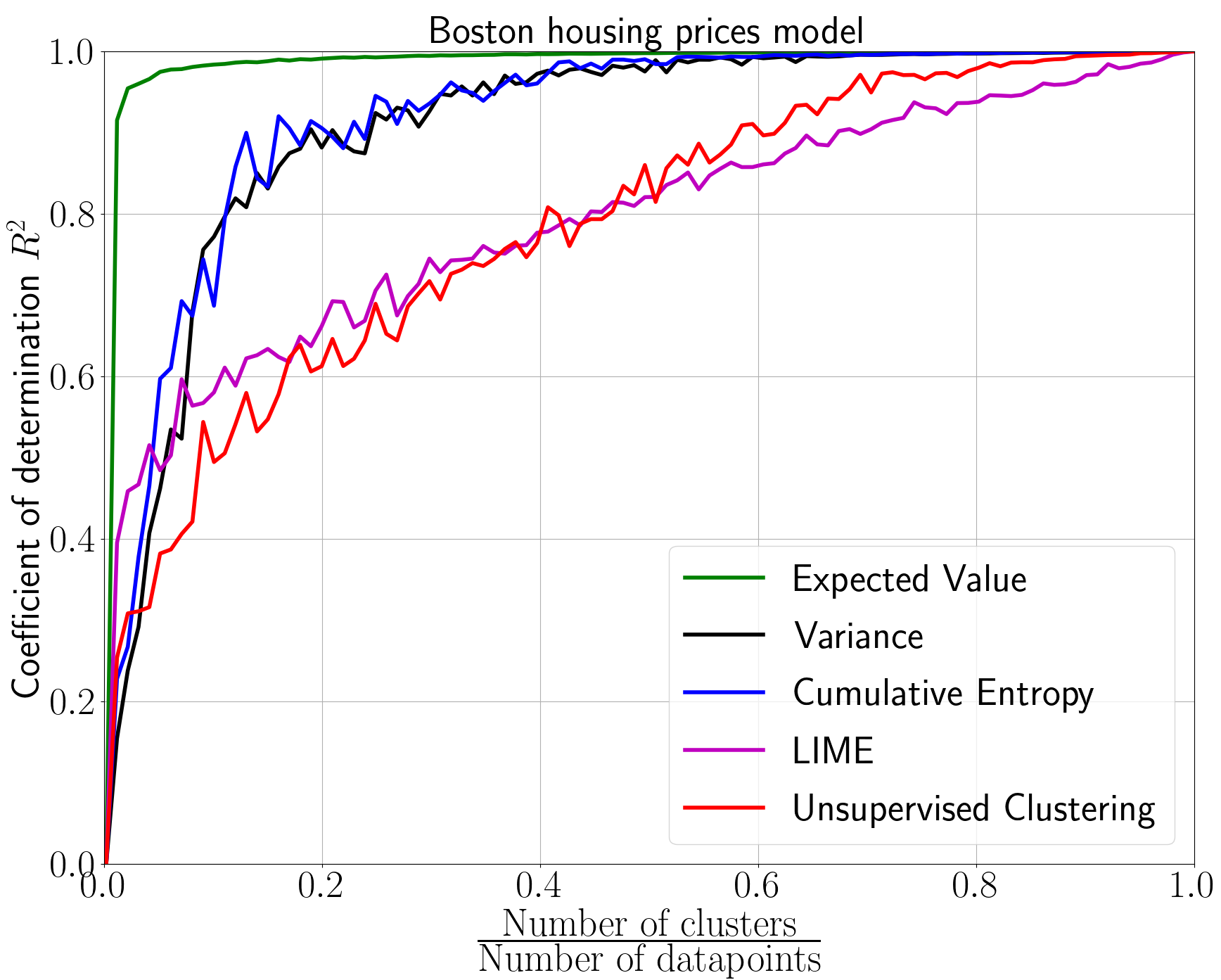}
\caption{\textbf{Supervised clustering performance} of different measures of interpretation for the random forest regressor trained on the Boston housing prices data set same as in \Figureref{supervised_boston_housing_loss}. We leave out Shannon entropy $H$ as the model output is continuous valued. LIME and unsupervised clustering result in inferior clusters. The measures based on statistical dispersion demonstrate better and comparable performance. The measure based on the expected value (KernelSHAP) demonstrates the best clustering which is not unexpected since the characteristic function is directly related to the model output $\bbE\left[g(\widetilde{\bX})|do(\widetilde{\bX}_S=x_S)\right]-\bbE[g(\bX)]$. }
\label{fig:supervised_boston_housing_model}
\end{figure}

\begin{figure}[h!]
\centering
 \includegraphics[width=0.7\linewidth]{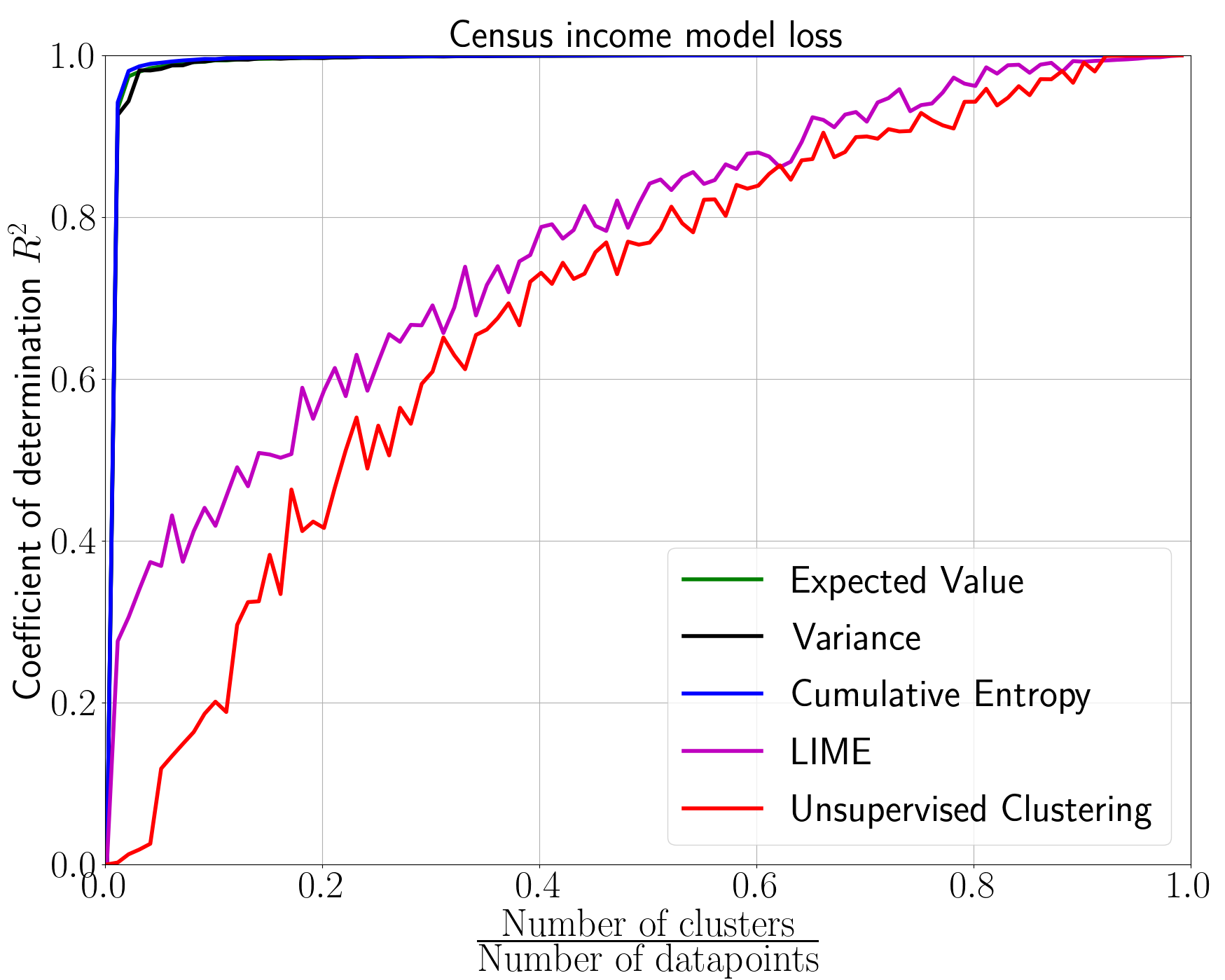}
\caption{\textbf{Supervised clustering for model loss} Analogous to \Figureref{supervised_boston_housing_loss}, we examine the quality of clusters obtained based on interpretations. We do not compare against interpretations using Shannon entropy $H$ because the mean squared error loss $l$ is continuous-valued. In this scenario, the $R^2$ corresponds to the percentage of the variance of the loss that is explained by the clustering which is performed in the space of interpretations which are $p+1$ dimensional including that for the target $Y$. Here we see improved performance in terms of the minimum number of clusters required for a given $R^2$ for all methods using interpretations, as compared to unsupervised clustering as well as LIME. The expected value measure as well as both cumulative paired entropy and variance which measure statistical dispersion demonstrate the best performance.}
\label{fig:supervised_census_loss}
\end{figure}

\begin{figure*}[h!]
\centering
\begin{subfigure}{0.5\textwidth}
  \centering
  \includegraphics[width=0.8\linewidth]{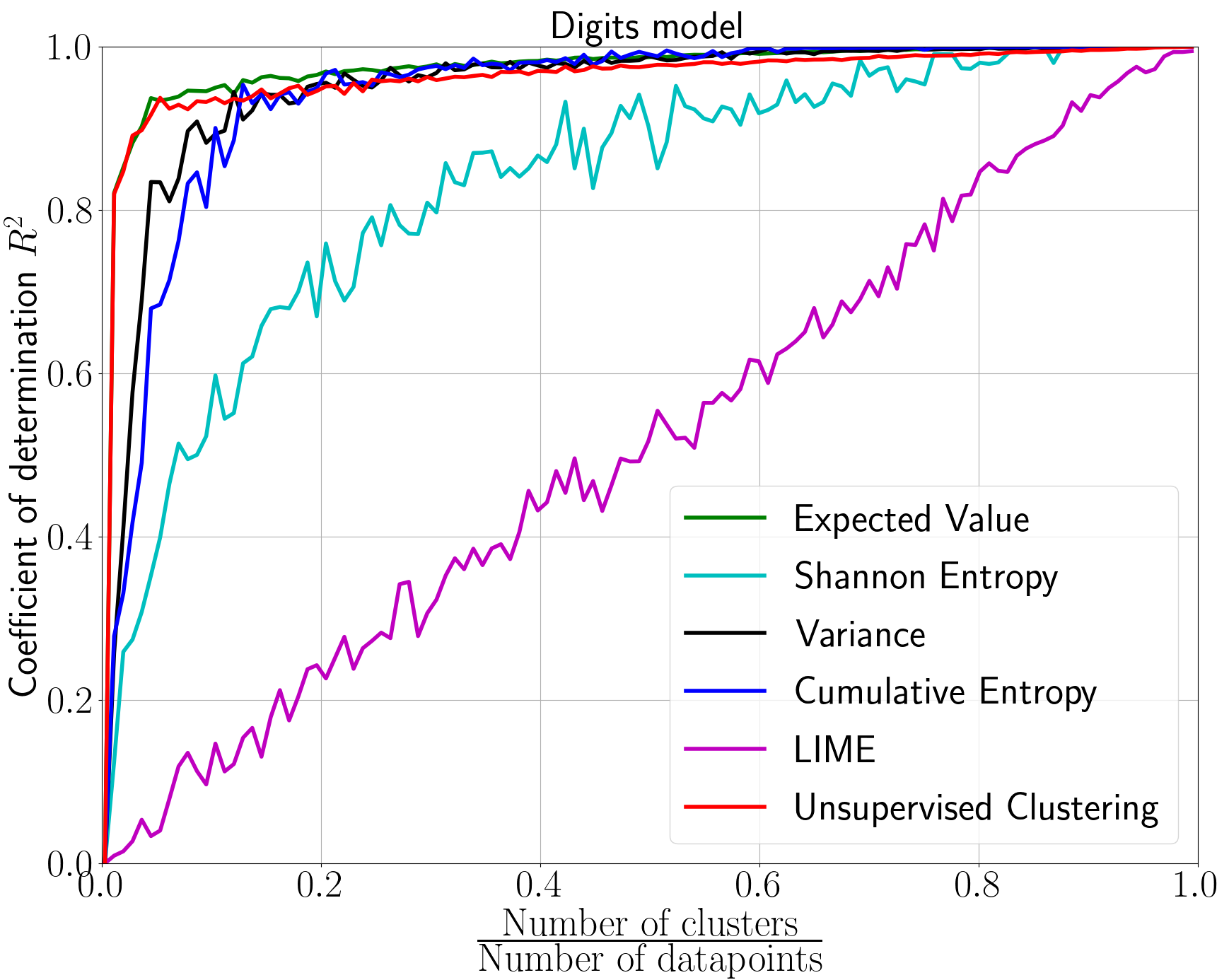}
  \label{fig:digits1}
\end{subfigure}\\
\vspace{0.5cm}
\begin{subfigure}{0.5\textwidth}
  \centering
  \includegraphics[width=0.8\linewidth]{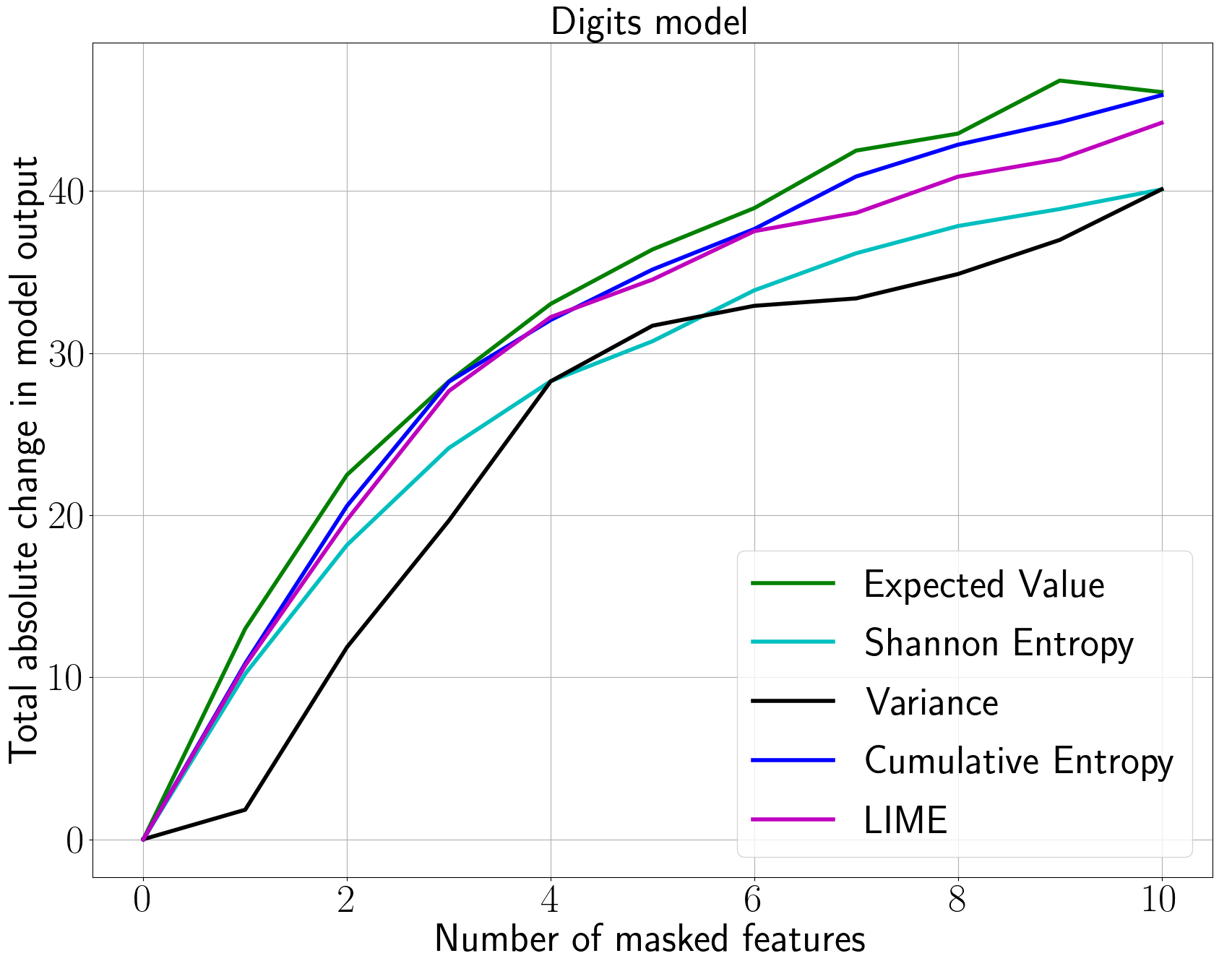}
  \label{fig:digits2}
\end{subfigure}\\
\vspace{0.5cm}
\begin{subfigure}{0.5\textwidth}
  \centering
  \includegraphics[width=0.8\linewidth]{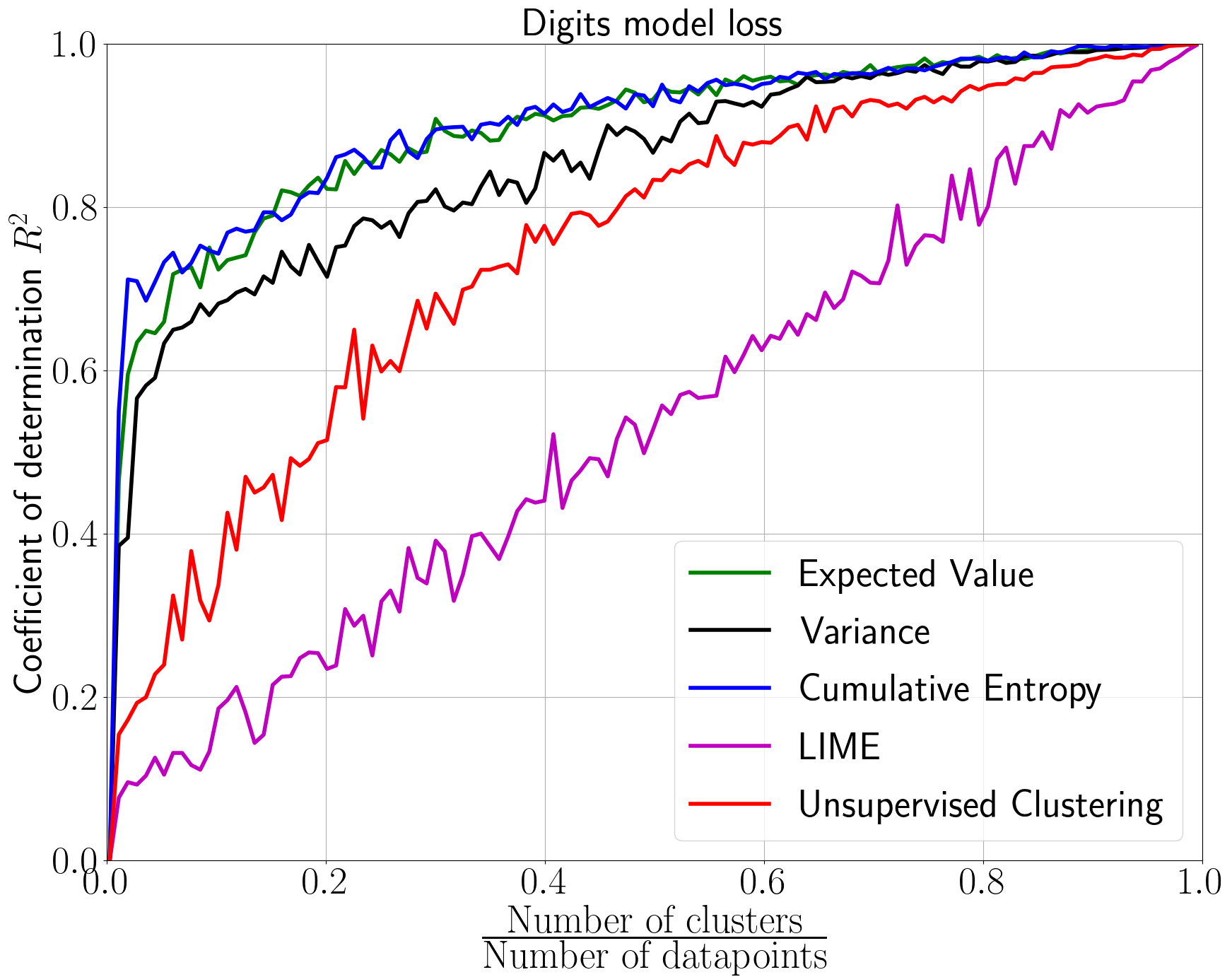}
  \label{fig:digits3}
\end{subfigure}
\caption{We perform analogous experiments as done in \Sectionref{results} for an SVM classifier trained on the handwritten digits data set \cite{uci} for classifying between 4 and 5. On evaluating the supervised clustering of interpretations for the model (top) we find all methods except LIME requiring a small number of clusters for achieving a high $R^2$. Unsupervised clustering is particularly effective here which is partially attributed to the features all being pixels. 
The same experiment is performed for interpretations about the cross entropy loss (bottom) in which we find the cumulative entropy \E based measure outperforming all others, with the next best being the expected value and the variance. Once again we find LIME to be ineffective in identifying good quality clusters. 
LIME is however able to perform comparably to the other methods in ordering features by importance (middle) where it is only outperformed by the expected value and cumulative entropy based measures of interpretation.} 
\label{fig:digits}
\end{figure*}
\end{document}